# Class Visualizations and Activation Atlases for Enhancing Interpretability in Deep Learning-Based Computational Pathology

Marco Gustav* (1), Fabian Wolf* (1), Christina Glasner* (2), Nic G. Reitsam (1,3,4), Stefan Schulz (2), Kira Aschenbroich (2), Bruno Märkl (3,4), Sebastian Foersch[#] (2), Jakob Nikolas Kather[#+] (1,5,6,7)

* equal first authorship

# equal last authorship

1. Else Kroener Fresenius Center for Digital Health, Faculty of Medicine, TUD Dresden University of Technology, Dresden, Germany
2. Institute of Pathology, University Medical Center Mainz, Mainz, Germany
3. Pathology, Faculty of Medicine, University of Augsburg, Augsburg, Germany
4. Bavarian Cancer Research Center (BZKF), Augsburg, Germany
5. Department of Medicine I, Faculty of Medicine, TUD Dresden University of Technology, Dresden, Germany
6. Medical Oncology, National Center for Tumor Diseases (NCT), University Hospital Heidelberg, Heidelberg, Germany
7. Pathology & Data Analytics, Leeds Institute of Medical Research at St James's, University of Leeds, Leeds, United Kingdom

+ Correspondence to:

Jakob Nikolas Kather, MD, MSc

Professor of Clinical Artificial Intelligence

Else Kroener Fresenius Center for Digital Health

Technical University Dresden

Fetscherstrasse 74

01307 Dresden, Germany

kather.jn@tu-dresden.de

## Abstract

The rapid adoption of transformer-based models in computational pathology has enabled the accurate prediction of molecular and clinical biomarkers directly from hematoxylin and eosin-stained whole-slide-images. However, explainability approaches have not kept pace with model complexity. While generative and attribution-based approaches are commonly used, feature visualization methods, including class visualizations (CVs) and activation atlases (AAs), have not been systematically evaluated for transformer-based histopathology models. We developed a concept-level feature visualization framework and evaluated CVs and AAs for a transformer-based foundation model in colorectal tissue and multi-organ cancer classification tasks with increasing label granularity. Four pathologists annotated real and generated images to quantify inter-observer agreement, complemented by attribution- and similarity-based metrics.

We found CVs preserved recognizability for morphologically distinct tissues but showed reduced separability for overlapping cancer subclasses, such as colon adenocarcinoma and rectum adenocarcinoma. In the tissue classification task, agreement decreased from Fleiss' $\kappa = 0.75$ for scans to $\kappa = 0.31$ for CVs; the cancer subclass tasks showed similar patterns at a lower baseline. AAs revealed layer-dependent organization; coarse tissue-level concepts formed coherent, well-separated regions, whereas finer subclasses exhibited dispersion and overlap. Agreement was moderate for tissue classification ($\kappa = 0.58$), high for coarse cancer groupings ($\kappa = 0.82$), and low at subclass level ($\kappa = 0.11$). Visual detail increased with network depth, reflecting progressive specialization. Atlas separability closely tracked expert agreement on real images, indicating that representational ambiguity reflected intrinsic pathological complexity and the abstraction level of intermediate representations. Attribution-based assignments approximated inter-pathologist variability in low-complexity settings, whereas perceptual and distributional metrics showed limited alignment.

Overall, concept-level feature visualization exposes structured morphological manifolds in transformer-based pathology models, reveals their organization, and mirrors expert ambiguity across label granularities, providing a framework for expert-centered interrogation of learned representations.

# Introduction

In computational pathology, deep learning (DL) models are used to infer clinically relevant patient-associated features from hematoxylin and eosin (H&E)-stained whole-slide images (WSIs). Weakly supervised learning strategies allow models to learn from slide-level labels without exhaustive region annotations, supporting prediction of molecular biomarkers[1–12], survival[12–15], and treatment response[12,16–18]. Contemporary pipelines typically extract patch-level feature representations using large pre-trained pathology-specific foundation models and aggregate these representations to obtain slide- or patient-level predictions.[9,19–21] Transformer-based architectures have further advanced this paradigm by modeling contextual interactions across image regions, improving performance across diverse computational pathology tasks.[9,19,20,22]

Despite these advances in predictive modeling, the internal organization of learned representations and the morphological patterns driving model predictions remain poorly understood. In computational pathology, explainable artificial intelligence (XAI) methods are predominantly instance-level and focus on explaining individual predictions (explainability), including saliency maps, gradient-based attribution, patch ranking, and low-dimensional embedding visualizations such as t-distributed stochastic neighbor embedding (t-SNE) or Uniform Manifold Approximation and Projection (UMAP).[9,10,23–26] Although these approaches localize image regions associated with predictions, they do not systematically characterize which morphological features define classes, are shared across classes, or constitute higher-order model-derived concepts. In contrast, representation-level interpretability focuses on how models organize and encode concepts within internal feature spaces rather than on explaining specific decisions.[26] A systematic analysis of representation structure is therefore required to characterize class-specific and shared features at the concept level, being essential for responsible deployment, as trust in DL-based systems depends on alignment between internal representations and established pathological reasoning.[24,27] Moreover, systematic interrogation may uncover prediction-relevant morphological patterns warranting further pathological investigation.

XAI approaches in computer vision can be broadly categorized into generative methods, attribution methods, and feature visualization techniques.[28–30] Generative approaches, including generative adversarial networks (GANs) and diffusion models, reconstruct visually realistic images from latent variables and can illustrate aspects of a model's internal feature space, but introduce additional modeling components and complexity.[30,31] Attribution methods compute and project predic-tion specific importance scores onto input pixels without modifying the model and are widely applied in computational pathology.[28,32–34] However, they remain inherently sample-specific and do not directly reveal class-level or slide-agnostic morphological concepts, nor do they characterize continuity and overlap between entities.

Feature visualization methods offer a complementary, model-centric perspective by directly exposing patterns encoded within trained networks.[28,35–37] Class visualization (CV) generates synthetic inputs that maximize class-specific activations, revealing prototypical patterns associated with individual prediction classes. Activation atlases (AAs), in contrast, organize internal activations into low-dimensional space and reconstruct representative feature patterns via feature inversion.[28,35,38] Particularly in convolutional neural networks (CNNs) trained on non-medical images, AAs have revealed coherent concept organization and transitions between visual features.[28,35–37] In computational histopathology, however, such approaches have only been marginally explored and have not been systematically evaluated for modern transformer-based foundation models.[20,22,39] Because Vision Transformers (ViTs) encode spatial and contextual information differently from CNNs, interpretability techniques developed for these architectures do not transfer directly.[28,35,38] It remains unknown whether transformer-based pathology models organize morphological concepts into structured internal representations that reflect pathological differences and yield features consistently identifiable by expert pathologists.

To address this, we developed a concept-level visualization framework and applied it to a ViT-based classification model trained on H&E-stained WSIs. We generated CVs and layer-wise AAs to reconstruct and systematically analyze internal feature representations across network depth and tissue classes. Analyses were conducted on nine colorectal tissue classes from the NCT-CRC-HE-100K dataset[40] and on multiple solid cancer classification tasks derived from The Cancer Genome Atlas (TCGA)[41,42], spanning increasing levels of label granularity and morphological overlap. We evaluated whether the organization of activation atlases reflects pathological distinctness and whether morphologically similar entities exhibit increased overlap. To this end, four pathologists independently annotated real and generated images to quantify interobserver agreement, and we implemented quantitative surrogate metrics, including attribution based, perceptual similarity, and distributional distance measures, to assess concordance with expert interpretation. By relating atlas separability to expert agreement across classification settings, we established a systematic framework to evaluate representation structure across tissue and cancer classification tasks.

# Materials and Methods

## Datasets

In this study we analyzed model activations on two morphologically diverse H&E datasets (Fig. 1). The NCT-CRC-HE-100K dataset (Tab. 1, Fig. 2A), hereafter NCT dataset, has been widely adopted in prior computational pathology studies[40] and comprises 100,000 non overlapping 224 × 224 pixels (px) patches (0.5 µm/px) from 86 colorectal cancer slides, manually annotated into nine tissue classes including tumor and non-tumor regions (Tab. 2, Fig. 2A). Images were color normalized using Macenko's method.[43] The second dataset, derived from The Cancer Genome Atlas[41,42] (Tab. 1, Fig. 2B, dataset publicly available under https://www.cancer.gov/tcga), hereafter referred to as TCGA, contains 1,608,060 patches randomly sampled from annotated tumor regions across 32 cancer types, of which eleven were selected to represent clinically relevant and morphologically diverse entities (Tab. 2, Fig. 2B). Patches were extracted at 0.5 µm/px with a size of 256 × 256 px without color normalization.

## Image Processing and Deep Learning Techniques

Feature visualization involves generating images that activate specific features within a defined layer in a neural network; CV, as a subset of this, is focusing on maximizing the activation in the final layer relevant to predicting a particular tissue or cancer class. Generated feature visualizations depend on the network architecture, learned parameters, selected layer, and optimization hyperparameters, and, for AAs, on the dataset from which target activations are derived. They should therefore be interpreted as model- and dataset-specific conceptual representations rather than faithful reconstructions of internal decision mechanisms. For training and activation capture, we employed the ViT-based foundation model UNI as backbone[20,38] with a subsequent linear classification layer that outputs class logits. The final linear layer was trained using a cross-entropy loss function, with frozen backbone weights. For consistency, the same model architecture was applied across all experiments, varying only the training dataset.

In detail, TCGA patches were randomly cropped to 224 × 224 px to match the input dimensions of the ViT architecture. A unified preprocessing pipeline was applied, including random flipping, rotation, and color jitter (brightness, contrast, and saturation), followed by normalization using ImageNet statistics (mean [0.485, 0.456, 0.406], standard deviation [0.229, 0.224, 0.225]). The NCT dataset was partitioned into five folds for cross validation, using four folds for training and one for validation in each iteration. The CRC-VAL-HE-7K dataset[40] was reserved for external testing. TCGA patches were grouped into six folds using StratifiedGroupKFold[44] to prevent patch level data leakage; one fold was held out for testing and the remaining five were used for cross validation. Models were trained using the AdamW optimizer[45] with a learning rate of 0.001 for up to 50 epochs, with early stopping after 20 epochs without improvement in validation loss. Although five-fold cross validation was used to estimate performance, feature and class visualizations were generated from a single model to limit computational cost, specifically the checkpoint with the lowest validation loss from the first training fold.

## Visualization Techniques

We adapted CVs and AAs for histopathology to expose model relevant morphological concepts rather than provide mechanistic explanations of model internals.[28,35,36] Both methods optimize an input image to induce predefined target activations (Fig. 1), either by maximizing a class logit (CV) or by matching previously captured and aggregated activations (AA). The selected network layer determines feature complexity, with earlier layers encoding simple structures and deeper layers integrating more complex patterns.[28,36] The target activation vector *f* may correspond to the full layer output or a subset, such as a specific token (e.g., the class token (cls)) or class logit. In contrast to model training, parameters remain fixed and the input image is iteratively optimized. Images were initialized randomly and parameterized in Fourier space to reduce high frequency artifacts and improve convergence, followed by inverse Fourier transform to obtain spatial domain visualizations.

**Class Visualization (CV)** generates class-specific feature representations by optimizing an input image to maximize a target class logit[35] (Fig. 1B). For each class *c* in the NCT and TCGA datasets, we optimized an image *img* such that $img^* = arg\ max_{img} cl(p, c)$, where $cl(p, c)$ denotes the pre-softmax logit for class $c$ under model parameters $p$.[22] Optimization of the raw class logits produced the most coherent and recognizable images and was performed via backpropagation for 8192 iterations, after which convergence was consistently observed.

Feature inversion was used to generate **Activation Atlases (AA)** by reconstructing randomly initialized input images from internal activations, enabling analysis of concept organization within the model's representations rather than individual predictions.[35] Activations corresponding to a predefined target feature vector $f^*$ at layer *l* of the trained transformer with parameters $p$ were first extracted for all training patches. These high-dimensional activations were embedded into two dimensions using t-distributed stochastic neighbor embedding (t-SNE)[46] (Fig. 1D–E) and partitioned the embedding into grids of 10 × 10 for NCT and 20 × 20 for TCGA, respectively. Within each cell, original activations were averaged component wise to aggregated target vectors. For each cell, an image *img** was optimized to match its target activation $f^*$ by minimizing $img^* = arg\ min_{img}\ \mathcal{L}(f_l(img;p), f^*)$[36,47], where $f_l(img;p)$ denotes the activation at layer *l*. We used the Euclidean (L2) loss $\mathcal{L}(f_l(img;p), f^*) = \|f_l(img;p) - f^*)\|_2^2$ and performed optimization via backpropagation for 8192 iterations until convergence. Optimized images were arranged according to their grid position to form the atlas.

For each cell, we additionally recorded the class distribution of underlying ground-truth (GT) patches and computed class attribution scores.[35] For a given class *c*, the attribution was defined as the mean dot product

$$\frac{\Sigma_i < f_i^*, \nabla_{f_i^*}\ cl(p,c) >}{n},$$

where <·,·> denotes the dot product, $f_i^*$ represents one of the cell's *n* individual target activations prior to averaging, and $\nabla_{f_i^*}\ cl(p,c)$ is the gradient of the class logit with respect to that activation. Thus, each cell's attribution score corresponds to the mean contribution of its unaggregated activations to the specified class logit. AAs and associated metadata can be visualized via the interactive visualization tool provided with this study (Tab. 1). Feature visualization was performed by extracting activations from selected layers of trained models using 224 × 224 px input patches without augmentation. Unless otherwise specified, activations were obtained from an intermediate layer (layer 14 of 24) of the UNI model for AA generation, as earlier layers primarily encoded low-level features, while deeper layers produced increasingly specialized and fragmented representations; for one model, all 24 layers were analyzed for comparison. In total, four models were trained: one on the NCT dataset including all nine tissue classes, and three on TCGA (Tab. 2) using clinically defined subsets. These comprised TCGA-5, including sarcoma (SARC), diffuse large B cell lymphoma (DLBCL), colorectal adenocarcinoma (COAD/READ), and melanoma (SKCM); TCGA-8, including lung squamous cell carcinoma (LUSC), lung adenocarcinoma (LUAD), breast carcinoma (BRCA), prostate adenocarcinoma (PRAD), colorectal adenocarcinoma (COAD/READ), kidney renal clear cell carcinoma (KIRC), and kidney renal papillary cell carcinoma (KIRP); and TCGA-11, combining all classes from TCGA-5 and TCGA-8. CVs were generated for the NCT (9 classes) and TCGA-11 (11 classes) models.

## Quantification of Informative Value in Generated Images

The generated images visualized model-relevant features, enabling pathologists to assess correspondence with established morphological patterns and to identify potentially unrecognized diagnostic features. Because concept-level visualizations lacked a single objective GT, we evaluated alignment with inter-observer agreement rather than predictive accuracy or causal faithfulness. We assessed agreement and classification performance on real and generated samples. A generated image's GT label was defined as the class it was optimized to visualize (CV) or as the majority GT class of the real samples from which the target activation was aggregated (AA). We then tested whether class attribution and quantitative metrics could serve as surrogates for pathologists' assessments by measuring their agreement with pathologist classifications and GT labels.

Conventional metrics such as mean squared error (MSE) and structural similarity index measure (SSIM) do not capture the context-dependent complexity of human visual perception.[48–50] Perceptual similarity is difficult to formalize because human judgments do not strictly satisfy mathematical properties of a distance metric.[48] We therefore used DL-based feature embeddings (e.g., UNI[20] for H&E images) to approximate perceptual similarity, particularly for complex image generation.[50] Except for attributions, all applied metrics were distance measures computed on features extracted from real and generated images, including Mahalanobis distance, learned perceptual image patch similarity (LPIPS), and DreamSim.

Mahalanobis Distance (MD) measures a distance between a point and a distribution of such points, assuming a class-conditional Gaussian distribution.[51–54] The MD of a point $P \in \mathbb{R}^N$ from a distribution with mean $\mu \in \mathbb{R}^N$ and covariance matrix $\Sigma \in \mathbb{R}^{N \times N}$ is defined as $\sqrt{(P-\mu)\Sigma^{-1}(P-\mu)}$. Unlike Euclidean distance (ED), MD accounts for feature correlations, which improves performance in high-dimensional spaces.[53,54] For numerical stability, we estimated covariance matrices using Ledoit-Wolf shrinkage[55] and computed their pseudo-inverse. We computed squared MDs between feature vectors of generated images and the class-specific distribution of sampled real-image features. Features were extracted using UNI, H-optimus-0[56], and Prov-GigaPath[57]. Lower MD indicated greater class consistency from the respective feature space. Due to memory constraints, we used only the final feature representations.

Learned Perceptual Image Patch Similarity (LPIPS) measures a distance between two images $img_1$, $img_2$ based on features extracted from multiple layers of a pre-trained CNN.[48,50] A lower LPIPS value corresponds to higher image similarity. The LPIPS value is calculated as

$$LPIPS(img_1, img_2) = \sum_l \frac{1}{H_l W_l} \sum_{h,w} \| w_l \odot (\hat{f}_{1,l}^{h,w} - \hat{f}_{2,l}^{h,w}) \|_2^2$$

with $\hat{f}_{1,l}^{h,w}$, $\hat{f}_{2,l}^{h,w} \in \mathbb{R}^{H_l \times W_l \times C_l}$ being features from layer *l* at spatial position (*h*, *w*) that have been unit-normalized across all $C_l$ channels. $w_l \in \mathbb{R}^{C_l}$ denotes channel-wise weights trained for image similarity. For transformer-based architectures, we replaced spatial dimensions ($H_l W_l$) with the token dimension and indexed positions by token. For linear layers, we set $H_l W_l = 1$. LPIPS distances were computed between generated images and real images for each GT class using AlexNet[58], UNI, H-Optimus-0, and Prov-GigaPath. For AlexNet, we adopted the layer selection and pretrained perceptual weights proposed by Zhang et al.[48] Although AlexNet was not trained on pathology images, the three pathology-specific networks were included to mitigate this limitation. For the transformer-based models, we did not train the layer-wise weights $w_l$ and fixed them to 1 (i.e., identity weighting).

DreamSim[50] is a learned perceptual metric that computes image distance from concatenated embeddings of DINO[59], CLIP[60] and OpenCLIP[61]. The ensemble was fine-tuned on mid-level variations (e.g., layout, pose, semantic content) to better align with human perception than metrics such as LPIPS. Given two images $img_1$ and $img_2$ with feature vectors $\hat{f}_1, \hat{f}_2 \in \mathbb{R}^C$, the distance is defined as

$$DreamSim(img_1, img_2) = 1 - \frac{< \hat{f}_1, \hat{f}_2 >}{\| \hat{f}_1 \|_2 \| \hat{f}_2 \|_2}$$

i.e., as the cosine distance between the two feature vectors. A lower DreamSim value indicates higher image similarity. As fine-tuning DreamSim on pathology images was beyond the scope of this study, we used the pretrained model to compute distances between generated images and real images for each GT class.

## Statistical Analysis

Statistical analyses quantified whether expert annotations of real and generated images reflected shared, human-recognizable morphological concepts rather than diagnostic accuracy or explanation correctness. Four pathologists, one of whom was board-certified, independently annotated images from the NCT tissue classification task (nine classes) and TCGA cancer classification tasks (up to eleven classes), comprising real images, CVs, and AA images. Annotators assigned a single class label or marked uncertainty; uncertain labels were either excluded or treated as a separate category, depending on the analysis. Inter-annotator agreement was assessed using Fleiss' κ (overall agreement), Cohen's κ (pairwise annotator and annotator-metric agreement), and Krippendorff's α (nominal, allowing missing labels). Agreement was interpreted as consistency of morphological interpretation, not model validation. For AA images, we computed agreement both between annotator pairs and between annotators and atlas-induced labels obtained using nearest-neighbor and distance-based feature aggregation. Uncertainty for Cohen's κ and percent agreement was estimated by bootstrap resampling of paired image-level labels (300 iterations, fixed seed); Krippendorff's α was computed on the full annotation matrix. Accuracy, F1 score, sensitivity, specificity, and confusion matrices were calculated descriptively against reference labels to assess class-discriminative information. These metrics were secondary, as CVs permit target-based assessment whereas AA images lack a single GT label and may support multiple valid interpretations. Analyses were conducted in Python 3.14.2 using scikit-learn 1.8.0, statsmodels 0.14.6, and krippendorff 0.8.2.

# Results

First, we trained a linear classifier on top of the UNI foundation model and achieved consistently high classification performance across datasets (Tab. 3). Based on this model, we then evaluated CVs on NCT and TCGA using performance and agreement metrics to assess class-specific recognizability and inter-annotator consistency, with scan-level annotations as baseline (Fig. 3–5). We then generated AAs to analyze internal representations and, in the absence of an objective GT, assessed inter-annotator agreement and quantitative surrogate metrics (Fig. 6–7).

### Class Visualizations Preserve Detectability of Multiple Tissue Classes

On NCT, scan annotations (Fig. 2A) showed high inter-annotator agreement and clear separability for most tissue classes, including adipose (ADI), lymphocytes (LYM), colorectal adenocarcinoma epithelium (TUM), smooth muscle (MUS), normal colon mucosa (NORM), and debris (DEB), with minor discrepancies (e.g mucus (MUC) vs. cancer-associated stroma (STR)), and infrequent use of the indeterminate label (Fig. 4). CVs largely preserved class-specific patterns (Fig. 3A) and maintained consistent detectability for several classes (ADI, LYM, NORM, TUM), capturing recognizable morphologies such as large lipid vacuoles in adipose tissue, dense lymphocytic infiltrates, and crowded pleomorphic glands in tumor epithelium; however, separability decreased and inter-annotator variability increased, particularly between STR and MUS (Fig. 4). Despite reduced agreement relative to scans, accuracy, F1 score, and sensitivity remained moderate to high, and specificity remained high (Fig. 5).

### Class Visualizations Show Reduced Separability and High Annotator Variability Across Multiple Solid Cancer Classes

In contrast, on TCGA-11, CVs exhibited substantially reduced separability and pronounced annotator-dependent variability (Fig. 3B). Although CVs reflected characteristic histopathological patterns across cancer entities, with adenocarcinomas from different organs (COAD, LUAD, PRAD) showing similar glandular architecture and other types displaying distinct features such as clear cell nesting in KIRC, spindle-cell fascicles in SARC, and diffuse lymphoid infiltrates in DLBCL, class discrimination remained feasible for selected entities but was reduced among morphologically overlapping classes. At the adenocarcinoma level (ADENO), dominant scan-level trends persisted (e.g., ADENO most frequent), yet frequent cross-assignments occurred (e.g., ADENO as LUSC, DLBCL, SARC, KIRC; SARC as ADENO), while ADENO and SKCM retained relatively higher consistency (Fig. 4). Accordingly, accuracy, F1 score, and sensitivity were lower than in NCT, and agreement was markedly reduced and more variable than for scan annotations (Fig. 5). At the subclass level, ambiguity further increased, with broadly distributed assignments and recurrent misclassifications (e.g., BRCA as KIRC, DLBCL as SARC, KIRC and SKCM as DLBCL; Fig. 4), resulting in uniformly low performance and minimal consensus for fine-grained CV interpretation (Fig. 5). Overall, NCT, comprising fewer visually distinct classes, showed high agreement and separability for scans and CVs with only moderate agreement loss, whereas TCGA exhibited reduced agreement and greater annotator-dependent variability, particularly at the subclass level, demonstrating that CV interpretability decreases with increasing dataset complexity and label granularity.

### Activation Atlases Capture Consistent Tissue-Level Concepts in the NCT dataset

Following the instance-level CV analysis, we next examined AAs to evaluate concept-level representations beyond individual logits. AA cells were independently annotated by four pathologists and compared with metric-derived labels; in the absence of an objective semantic GT, expert annotations served as reference. Across the 24 UNI transformer layers, AAs and embeddings showed increasing structural detail and class differentiation with depth for the NCT dataset (Fig. S1–S5). Early layers encoded low-level morphology with limited separability, intermediate layers exhibited coherent concept organization and stable clustering, and deeper layers produced more detailed but increasingly specialized representations. We therefore systematically inspected all layers and selected layer 14/24 as a balance between separability and over-specialization for subsequent analyses.

For NCT, inter-annotator agreement was moderate (Fleiss' $\kappa = 0.58$; Krippendorff's $\alpha = 0.56$) with comparable pairwise Cohen's κ across annotator pairs, but varied across classes and cells, which often received heterogeneous labels (Fig. 6–7). Attribution-based assignments achieved agreement within the inter-annotator range and reproduced class prevalence closely matching expert annotations without mode collapse. Among perceptual metrics, nearest-neighbor (NN)-based DreamSim and LPIPS outperformed distance-based variants; DreamSim reached higher κ than LPIPS but showed greater annotator-dependent variability (Fig. 7), while LPIPS with Prov-GigaPath overlapped attribution-level agreement in selected comparisons. DreamSim distributed prevalence across more classes, whereas LPIPS concentrated assignments depending on the feature extractor (Fig. 6). In contrast, MD showed near-zero agreement across extractors and collapsed assignments to few classes, with minimal overlap with expert labels.

## Annotator Agreement and Metric Alignment in Activation Atlases Vary with Label Complexity in TCGA Cancer Classes

Next, we analyzed AAs across TCGA class sets to assess the effect of label granularity (Fig. S6–S17). Agreement decreased with increasing class complexity: at the adenocarcinoma level, Fleiss' κ/Krippendorff's α were 0.82/0.82 (TCGA-5, Fig. S7), 0.30/0.30 (TCGA-8, Fig. S11), and 0.37/0.37 (TCGA-11, Fig. S15), with increasing dispersion of pairwise Cohen's κ and more heterogeneous prevalence maps (Fig. S6, S10, S14). Attribution-based labels overlapped inter-annotator agreement at coarser levels and distributed prevalence across classes without dominance. Among perceptual metrics, NN-DreamSim consistently outperformed distance-based variants and overlapped attribution and selected annotator agreement, albeit with higher variability. NN-LPIPS (Prov-GigaPath) achieved among the highest κ values, though rankings varied by class set and feature extractor; DreamSim tended to concentrate assignments on dominant classes, whereas LPIPS distributed them more broadly. At the subclass level, agreement was uniformly low (TCGA-5: $\kappa = 0.59$, $\alpha = 0.62$; TCGA-11: $\kappa = 0.11$, $\alpha = 0.24$), with increased annotator variability and reduced metric alignment. MD again showed near-zero agreement and collapsed assignments to few classes; uncertain labels were used by one annotator only at the adenocarcinoma level (Fig. S6–S17).

Overall, AAs captured tissue- and cancer-level concepts at coarse label resolutions, with agreement and metric alignment comparable to inter-observer variability; both declined with increasing label granularity, paralleling reduced expert consensus in real images and the patterns observed for instance-level CVs, indicating sensitivity to dataset complexity.

# Discussion

Concept-level feature visualization demonstrated that transformer-based pathology models organize morphological information into structured latent representations. CVs and AAs did not provide mechanistic explanations of individual predictions but revealed how concepts and their relationships were encoded. Coarse tissue categories yielded coherent, well-separated patterns, whereas increasing label granularity resulted in greater dispersion and overlap, indicating that separability scaled with morphological complexity and that transformer representations preserved intrinsic pathological structure rather than imposing artificial boundaries. Layer-wise analyses showed progressively increased structural detail and class differentiation with depth, consistent with prior CNN findings.[35,62]

At the level of individual classes, CVs recapitulated prototypic histomorphological features, such as dense aggregates of small hyperchromatic nuclei for lymphocytes and optically empty vacuoles within thin septa for adipose tissue. Reduced separability between STR and MUS in the NCT dataset, accompanied by increased inter-annotator variability, likely reflected their overlapping eosinophilic staining and shared spindle-cell morphology in colorectal tissue (e.g., collagen-rich stroma versus smooth muscle bundles). AAs further revealed continuity between related entities. In the NCT dataset, internal representations were compact and distinct, whereas morphologically overlapping TCGA cancer classes exhibited increased overlap and annotator-dependent ambiguity. Decreased atlas separability paralleled reduced expert agreement on real images, indicating that representational ambiguity reflected intrinsic pathological complexity rather than visualization artifacts.

Concept-level visualization complements attribution methods that localize relevance within individual images[24,25] and generative approaches relying on auxiliary models[30,31]. CVs identify patterns that maximize specific logits, whereas AAs expose density, continuity, and overlap of representations across layers, which is particularly relevant for transformer architectures, where dis-tributed representations and token interactions complicate interpretation through localized relevance maps alone.[28,35,39] The dependence on dataset complexity further indicates that representation structure is task-specific and not uniformly separable across granularities.

Quantitative surrogate metrics showed variable alignment with expert interpretation; attribution-based assignments achieved agreement comparable to inter-pathologist variability in settings with limited morphological overlap, suggesting that class attribution can approximate expert reasoning when class boundaries are distinct. Perceptual metrics (LPIPS, DreamSim) were informative but sensitive to feature extractor and task configuration. MD consistently failed to capture semantic structure, likely due to violations of class-conditional Gaussian assumptions in high-dimensional feature spaces with substantial intra-class variability.[53–55] No metric fully substituted expert assessment, as increasing morphological complexity reduced both metric alignment and inter-expert agreement, reflecting intrinsic diagnostic ambiguity rather than metric limitations alone.

Limitations include sensitivity of feature visualizations to architecture, layer selection (middle layers balanced separability and abstraction, whereas later layers showed more detailed reconstructions and higher similarity to real images), optimization strategy, and dimensionality reduction (e.g. t-SNE or UMAP).[28,29,35] Generated images reflect reconstructions within the training distribution and should be interpreted as exploratory summaries of learned patterns rather than causal explanations of decision mechanisms. Moreover, latent features remain abstract and interact non-linearly, risking overinterpretation.[47] Quantitative evaluation is further constrained by the absence of pathology-specific perceptual metrics; LPIPS and DreamSim were developed for natural images and emphasize global structure over subtle cellular morphology.[48,50] Domain-specific fine-tuning grounded in expert similarity judgments could improve metric validity.

Future work should extend this framework to narrowly defined differentiation tasks or molecular subtypes to enable validation against established genotype–phenotype correlations and identify residual, potentially underrecognized features.[11,25] Scaling from patches to WSIs would facilitate analysis of spatial context and multiscale organization, contingent on computational feasibility. Robustness should be assessed across image initializations and optimization settings, and systematic comparative analyses across layers, given their demonstrated influence on representational structure, and across architectures may clarify how morphological concepts emerge within representation hierarchies. Finally, development of pathology-specific perceptual metrics remains essential to better align quantitative similarity measures with expert judgment.

In conclusion, concept-level feature visualization provides a structured characterization of how transformer-based pathology models encode morphological information. By revealing overlap, continuity, and task-dependent separability within representation spaces, it reflects intrinsic morphological complexity and expert agreement across label granularities. This framework advances interpretability beyond single-prediction explanations by enabling systematic analysis of internal representation structure, thereby supporting transparent model evaluation, identifying limits of class separability, and facilitating informed human-AI interaction in computational pathology.

# Resource availability

## Lead contact

Further information and requests should be addressed to the lead contact, Jakob Nikolas Kather (kather.jn@tu-dresden.de).

## Data and code availability

All data and source code used in this study are publicly available. The NCT-CRC-HE-100K dataset[40] is available at https://zenodo.org/records/1214456. The TCGA-based cancer classification datasets derived from The Cancer Genome Atlas (TCGA)[41,42] are available at https://zenodo.org/records/5889558. All source code used to conduct this study is publicly available at https://github.com/KatherLab/PathoActivationAtlas with publication of the peer-reviewed paper, ensuring full reproducibility. Corresponding access links are summarized in Tab. 1.

# Ethics approval and consent to participate

This study was performed in accordance with the Declaration of Helsinki. This study is a retrospective analysis of scanned images of anonymized tissue samples of various cohorts of cancer patients. Data were collected and anonymized and ethical approval was obtained. The overall analysis was approved by the Ethics board of the Medical Faculty of Technical University Dresden under the ID BO-EK-444102022.

# Acknowledgements

The authors are grateful to the Center for Information Services and High Performance Computing (Zentrum für Informationsdienste und Hochleistungsrechnen (ZIH)) at TU Dresden for providing its facilities for high throughput calculations. Aspects of this work are part of the medical doctoral thesis of KA. JNK is supported by the German Cancer Aid DKH (DECADE, 70115166), the German Federal Ministry of Research, Technology and Space BMFTR (PEARL, 01KD2104C;

CAMINO, 01EO2101; TRANSFORM LIVER, 031L0312A; TANGERINE, 01KT2302 through ERA-NET Transcan; Come2Data, 16DKZ2044A; DEEP-HCC, 031L0315A; DECIPHER-M, 01KD2420A; NextBIG, 01ZU2402A; PROSURV, 01KD2509C), the German Research Foundation (DFG, Deutsche Forschungsgemeinschaft) as part of Germany's Excellence Strategy – EXC 2050/2 – Project ID 390696704 – Cluster of Excellence "Centre for Tactile Internet with Human-in-the-Loop" (CeTI) of Technische Universität Dresden, as well as through DFG-funded collaborative research projects (TRR 412/1, 535081457; SFB 1709/1 2025, 533056198), the German Academic Exchange Service DAAD (SECAI, 57616814), the German Federal Joint Committee G-BA (TransplantKI, 01VSF21048), the European Union EU's Horizon Europe research and innovation programme (ODELIA, 101057091; GENIAL, 101096312), the European Research Council ERC (NADIR, 101114631), the Breast Cancer Research Foundation (BELLADONNA, BCRF-25-225) and the National Institute for Health and Care Research NIHR (Leeds Biomedical Research Centre, NIHR203331). The views expressed are those of the author(s) and not necessarily those of the NHS, the NIHR or the Department of Health and Social Care. This work was funded by the European Union. Views and opinions expressed are, however, those of the author(s) only and do not necessarily reflect those of the European Union. Neither the European Union nor the granting authority can be held responsible for them. NGR is supported by the clinician scientist programme of the Faculty of Medicine, University of Augsburg, Germany. SF is supported by the Federal Ministry of Research, Technology and Space (DECIPHER-M, 01KD2420E), the German Cancer Aid (DECADE, 70115166 and TargHet, 70115995)

# Author contributions

MG, FW, NGR, SF, JNK conceptualized and designed the study. FW, CG developed the methodological framework for concept-level feature visualization. MG, FW curated and preprocessed the source datasets. FW implemented the deep learning training and inference pipeline. CG, FW developed the code for class visualizations and activation atlases. CG implemented quantitative evaluation metrics. MG, CG, FW developed the software for data analysis, visualization, and interactive exploration tools. MG, CG, FW planned and conducted the computational experiments. MG performed the analysis and interpretation of model outputs and visualization of results. NGR, SF, StS, KA, BM conducted pathological examination of samples and interpreted generated visualizations. MG, NGR, SF, JNK contributed to the interpretation of results in a clinical and methodological context. MG drafted the initial manuscript. MG prepared figures and supplementary materials. All authors revised the manuscript critically for important intellectual content, jointly interpreted the data, and approved the final version of the manuscript for submission.

# Declaration of interests

JNK declares ongoing consulting services for AstraZeneca and Bioptimus. Furthermore, he holds shares in StratifAI, Synagen, and Spira Labs, has received institutional research grants from GSK and AstraZeneca, as well as honoraria from AstraZeneca, Bayer, Daiichi Sankyo, Eisai, Janssen, Merck, MSD, BMS, Roche, Pfizer, and Fresenius. MG received honoraria for lectures sponsored by Sartorius AG, Germany and AstraZeneca, UK. SF received honoraria by AstraZeneca, BMS and MSD. NGR received travel support from Bruker Spatial Biology.

## Declaration of generative AI and AI-assisted technologies in the writing process

After preparing the initial manuscript, the authors used ChatGPT (OpenAI, GPT-5.2) and DeepL Write (DeepL SE) to edit selected sections to improve readability. After using these tools, the authors carefully reviewed and edited the text as needed and take full responsibility for the content of the publication.

# Figures

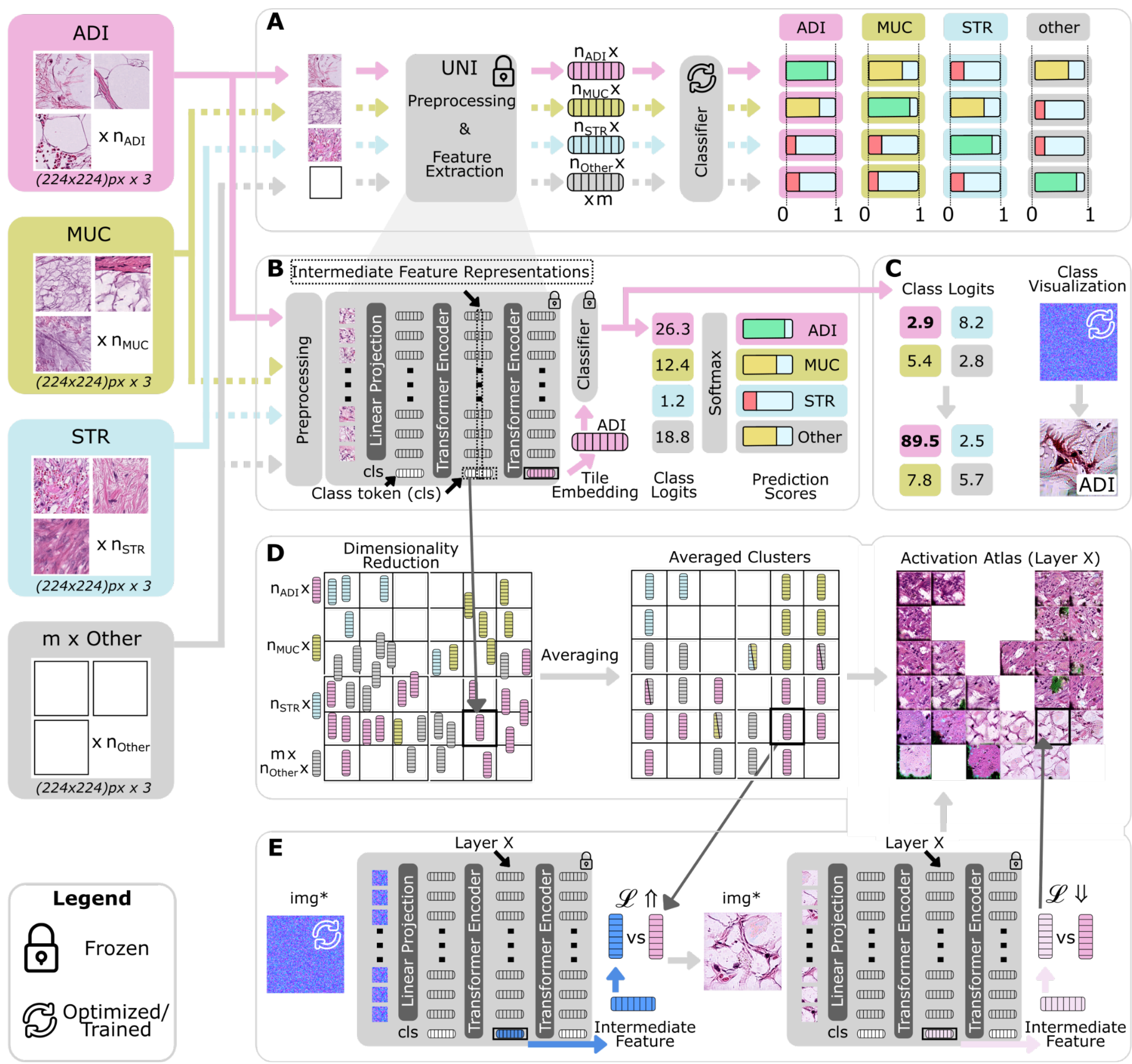

**Fig. 1 | Workflow for feature visualization in transformer-based computational pathology.** Overview of the proposed framework for analyzing transformer-based representations, including class visualizations and activation atlases for the National Center for Tumor Diseases (NCT) dataset. Logos in the legend were sourced from Canva (https://www.canva.com).

**A** Model training. Hematoxylin and eosin (H&E)–stained image patches from multiple tissue (NCT dataset) or cancer classes (TCGA dataset) are used to train a classifier consisting of a frozen vision transformer-based foundation model (UNI[20]) as feature extractor and a trainable linear classification head. The classifier outputs class logits for each patch.

**B** Inner workings and activation capturing. During inference, input patches are processed by the foundation model to produce token-level intermediate feature representations. Activations are captured from selected transformer layers of the trained model. Examples for potential activation capture are indicated by dotted circles. The cls token is passed to the linear classification head to produce class logits, which are optionally converted into class probabilities via softmax for prediction score visualization. In this study, we focused on the cls tokens from the respective layers for downstream analyses, as described in the methods section.
**C** Class visualization. Class visualizations are generated by optimizing a synthetic input image to maximize a specific class logit, producing a patch-level image that strongly activates the target class.
**D** Dimensionality reduction and averaging. High-dimensional patch-level activations are projected into a two-dimensional embedding space using dimensionality reduction (t-SNE). The embedding space is partitioned into a regular grid, and activations falling into the same grid cell are averaged to obtain representative target feature vectors for downstream visualization.
**E** Activation atlases. Feature inversion is applied to reconstruct images from target feature vectors captured in **B**. For individual target vectors, optimization minimizes the distance between the activation of the synthetic image and the target activation (illustrated for high- and low-loss ($\mathcal{L}$) cases). For activation atlases, optimized images are arranged according to their positions in the reduced embedding space, yielding a grid that visualizes concept-level morphological patterns learned by the model.

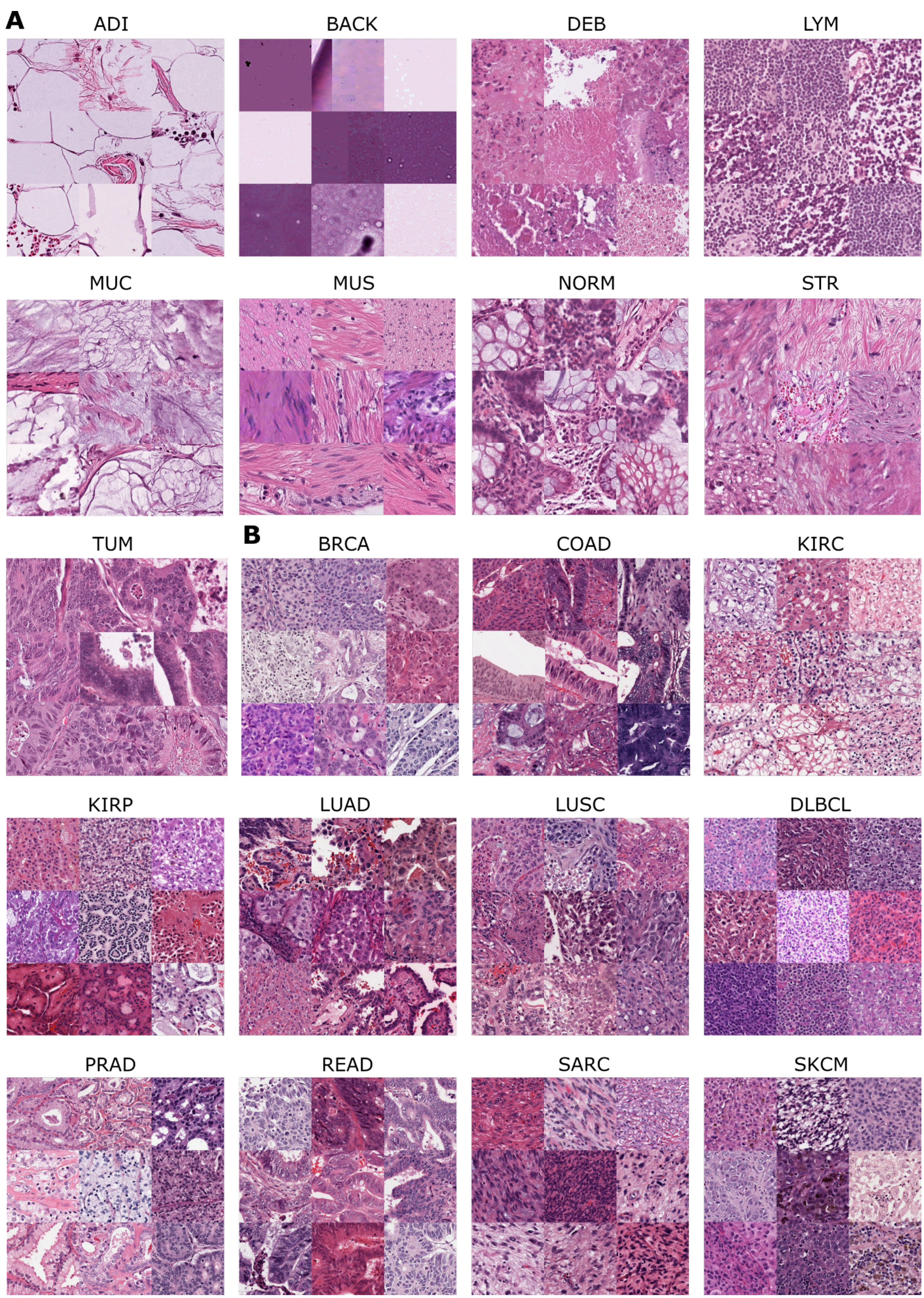

**Fig. 2 | Representative histopathology patches for both datasets.**

Each panel displays example patches illustrating characteristic morphological patterns associated with the corresponding tissue/cancer class as observed in whole-slide images.

**A** Representative hematoxylin and eosin (H&E)-stained image patches for the nine tissue classes used in the National Center for Tumor Diseases (NCT) dataset: adipose tissue (ADI), background (BACK), debris (DEB), lymphocytes (LYM), mucus (MUC), smooth muscle (MUS), normal colon mucosa (NORM), cancer-associated stroma (STR), and colorectal adenocarcinoma epithelium (TUM).

**B** Shown are representative H&E-stained image patches for the eleven cancer classes used from The Cancer Genome Atlas (TCGA) dataset: breast invasive carcinoma (BRCA), colon adenocarcinoma (COAD), kidney renal clear cell carcinoma (KIRC), kidney renal papillary cell carcinoma (KIRP), lung adenocarcinoma (LUAD), lung squamous cell carcinoma (LUSC), lymphoid neoplasm diffuse large B-cell lymphoma (DLBCL), prostate adenocarcinoma (PRAD), rectum adenocarcinoma (READ), sarcoma (SARC), and skin cutaneous melanoma (SKCM).

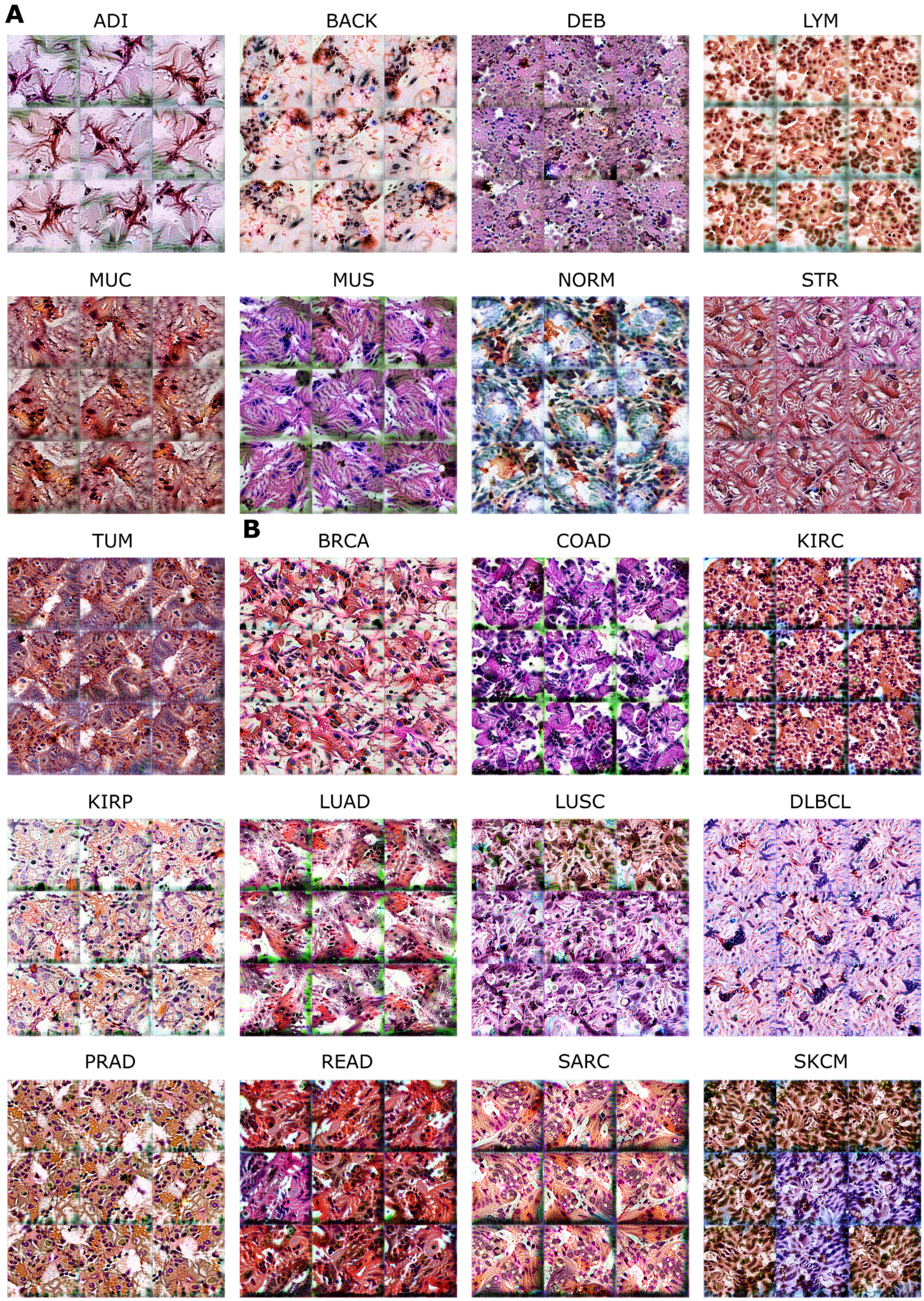

**Fig. 3 | Class visualizations for tissue classes for both datasets.**

For each class, synthetic images were generated by optimizing an input image to maximize the corresponding class logit of a trained vision transformer-based model (UNI[20]), while keeping all model parameters fixed. Each panel displays a grid of independently optimized class visualizations, illustrating morphological patterns that strongly activate the respective class according to the model. The visualizations are instance-agnostic and reflect class-level feature representations learned from hematoxylin and eosin (H&E)-stained image patches.

**A** Shown are class visualizations generated for the nine tissue classes of the National Center for Tumor Diseases (NCT) dataset: adipose tissue (ADI), background (BACK), debris (DEB), lymphocytes (LYM), mucus (MUC), smooth muscle (MUS), normal colon mucosa (NORM), cancer-associated stroma (STR), and colorectal adenocarcinoma epithelium (TUM). The CVs correspond to recognizable and well-characterized histopathological features, representing prototypic histopathological patterns: For adipose tissue, large 'empty' fat vacuoles with thin septa (classic fat morphology); for background, pale low-cellularity regions with minimal tissue structure (corresponding to slide background and processing artifacts); for debris, amorphous eosinophilic material with fragmented nuclei and necrotic remnants indicative of tissue breakdown; for lymphocytes, dense small nuclei with high cellularity; for mucus, extracellular mucin pools with sparse cellular elements; for muscle, elongated spindle-shaped cells arranged in bundles with eosinophilic cytoplasm; for normal, relatively organized architecture with epithelial cells; for stroma, fibrous collagen-rich areas; for tumor, crowded pleomorphic epithelial cells forming irregular glandular patterns (characteristic of CRC).

**B** Shown are class visualizations generated for eleven cancer classes from The Cancer Genome Atlas (TCGA): breast invasive carcinoma (BRCA), colon adenocarcinoma (COAD), kidney renal clear cell carcinoma (KIRC), kidney renal papillary cell carcinoma (KIRP), lung adenocarcinoma (LUAD), lung squamous cell carcinoma (LUSC), lymphoid neoplasm diffuse large B-cell lymphoma (DLBCL), prostate adenocarcinoma (PRAD), rectum adenocarcinoma (READ), sarcoma (SARC), and skin cutaneous melanoma (SKCM). The CVs correspond to prototypic histopathological patterns of the respective cancer types. For most adenocarcinomas (COAD, READ, LUAD, PRAD), the visualizations share common glandular architectural features reflecting their shared epithelial origin. For BRCA, irregular ductal or lobular arrangements of pleomorphic tumor cells embedded in desmoplastic stroma are prominent. For KIRC, tumor cells appear organized in nested or sheet-like configurations, consistent with the characteristic clear cell architecture, while KIRP visualizations suggest a papillary growth pattern. For LUSC, the visualizations resemble nests or sheets of polygonal squamous cells with features of keratinization. For SARC, spindle-shaped cellular arrangements in fascicular patterns reflect the mesenchymal origin of these tumors. For SKCM, a diffuse growth pattern with occasional melanin deposition is apparent. For DLBCL, diffuse sheets of large lymphoid cells with prominent nuclei are visible.

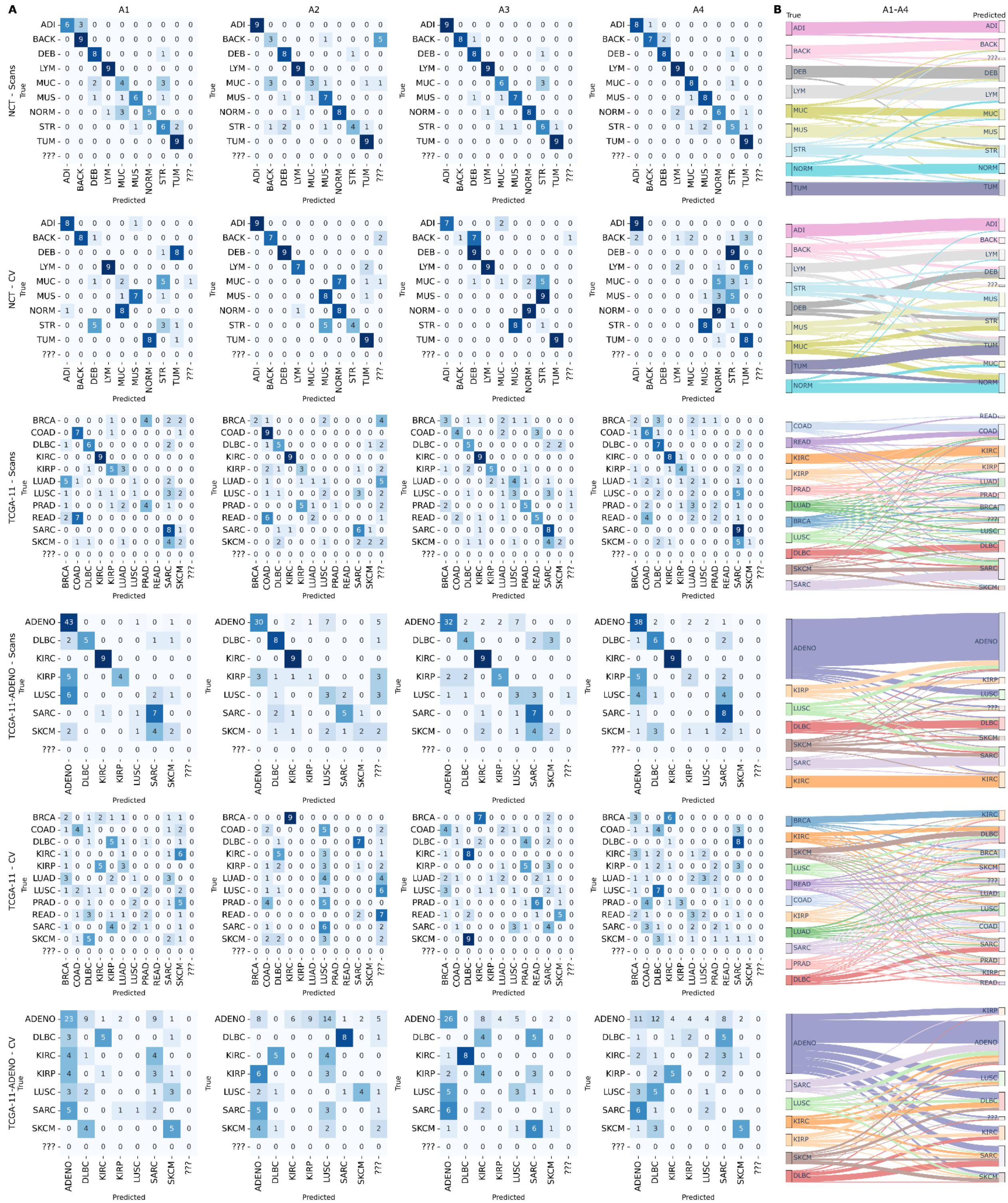

**Fig. 4 | Pathologist annotation patterns for scan images and class visualizations across tasks.**

We evaluated pathologist annotations for real scan images and model-derived class visualizations (CVs) across colorectal tissue classification (National Center for Tumor Diseases (NCT) dataset) and multi-cancer classification settings derived from The Cancer Genome Atlas (TCGA), including analyses at different levels of label granularity. The NCT task comprises nine colorectal tissue classes: adipose tissue (ADI), background (BACK), debris (DEB), lymphocytes (LYM), mucus (MUC), smooth muscle (MUS), normal colon mucosa (NORM), cancer-associated stroma (STR), and colorectal adenocarcinoma epithelium (TUM). TCGA-based settings comprise eleven solid tumor entities: breast invasive carcinoma (BRCA), colon adenocarcinoma (COAD), kidney renal clear cell carcinoma (KIRC), kidney renal papillary cell carcinoma (KIRP), lung adenocarcinoma (LUAD), lung squamous cell carcinoma (LUSC), lymphoid neoplasm diffuse large B-cell lymphoma (DLBCL), prostate adenocarcinoma (PRAD), rectum adenocarcinoma (READ), sarcoma (SARC), and skin cutaneous melanoma (SKCM). An additional adenocarcinoma-level grouping (ADENO) collapses morphologically related adenocarcinoma entities into a single class. An additional category ("???") denotes uncertain assignments.

**A** Row-normalized confusion matrices summarizing individual pathologist annotations (A1–A4) for scanned images and CVs. Rows denote reference labels (ground truth for scans; optimization objective class for CVs), and columns denote annotated labels. Color intensity represents the proportion of samples within each reference class assigned to a given label; overlaid numbers indicate raw annotation counts.

**B** Sankey diagrams showing the aggregated correspondence between reference labels (left) and pathologist-assigned labels (right), based on annotations summed across all pathologists. Ribbon widths are proportional to the number of annotated samples per label pair and depict class-wise agreement, systematic confusions, and label uncertainty.

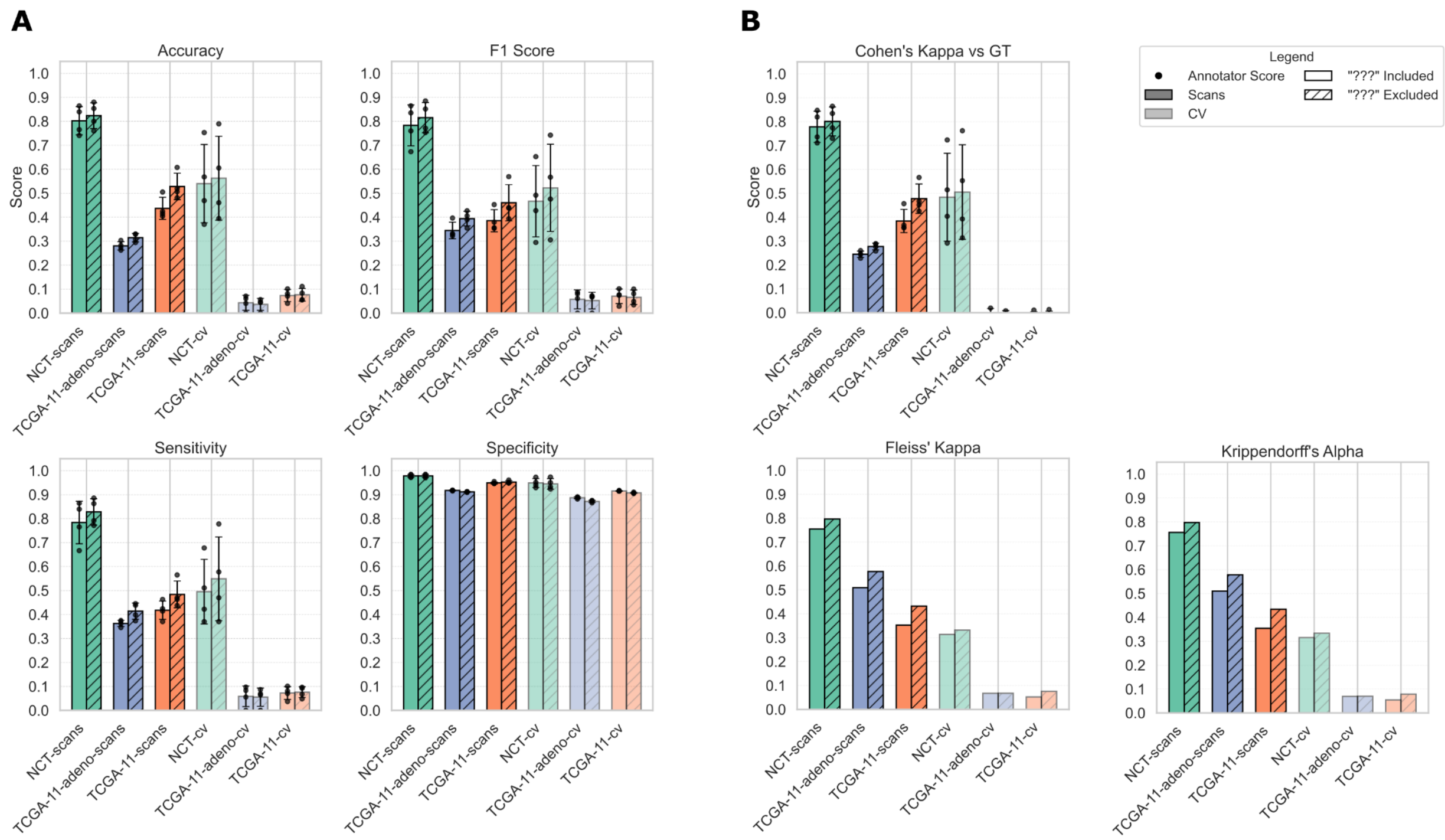

**Fig. 5 | Annotation performance and agreement metrics for class visualizations across datasets.**

Summary of pathologist annotation performance and agreement for real scan images and class visualizations (CVs) across the National Center for Tumor Diseases tissue dataset (NCT) and The Cancer Genome Atlas (TCGA)-derived classification tasks. Metrics are shown separately for real scan images (scans) and class visualizations (CVs), and for different label granularities, including the NCT tissue classification task, the TCGA-11 cancer classification task, and an adenocarcinoma-level grouping (TCGA-11-adeno), in which multiple adenocarcinoma entities are collapsed into a single class (ADENO). Hatched bars indicate analyses excluding uncertain annotations ("???"), whereas solid bars indicate analyses including uncertain labels as an explicit category. Metrics are reported to characterize annotation consistency and ambiguity rather than diagnostic accuracy or explanation correctness.

**A** Descriptive performance metrics including accuracy, F1 score, sensitivity, and specificity, computed with respect to reference labels. Bar heights indicate the mean of annotator-specific scores for each task and metric, error bars denote the standard deviation across annotators, and overlaid points represent individual annotator scores.

**B** Agreement metrics include Cohen's κ (agreement between individual annotators and reference labels), Fleiss' κ (overall agreement among multiple annotators), and Krippendorff's α, which accounts for missing labels. For Cohen's κ versus ground truth, bar heights indicate the mean across annotators, error bars denote the standard deviation, and overlaid points represent individual annotators. For Fleiss' κ and Krippendorff's α, bars indicate a single group-level agreement value without error bars.

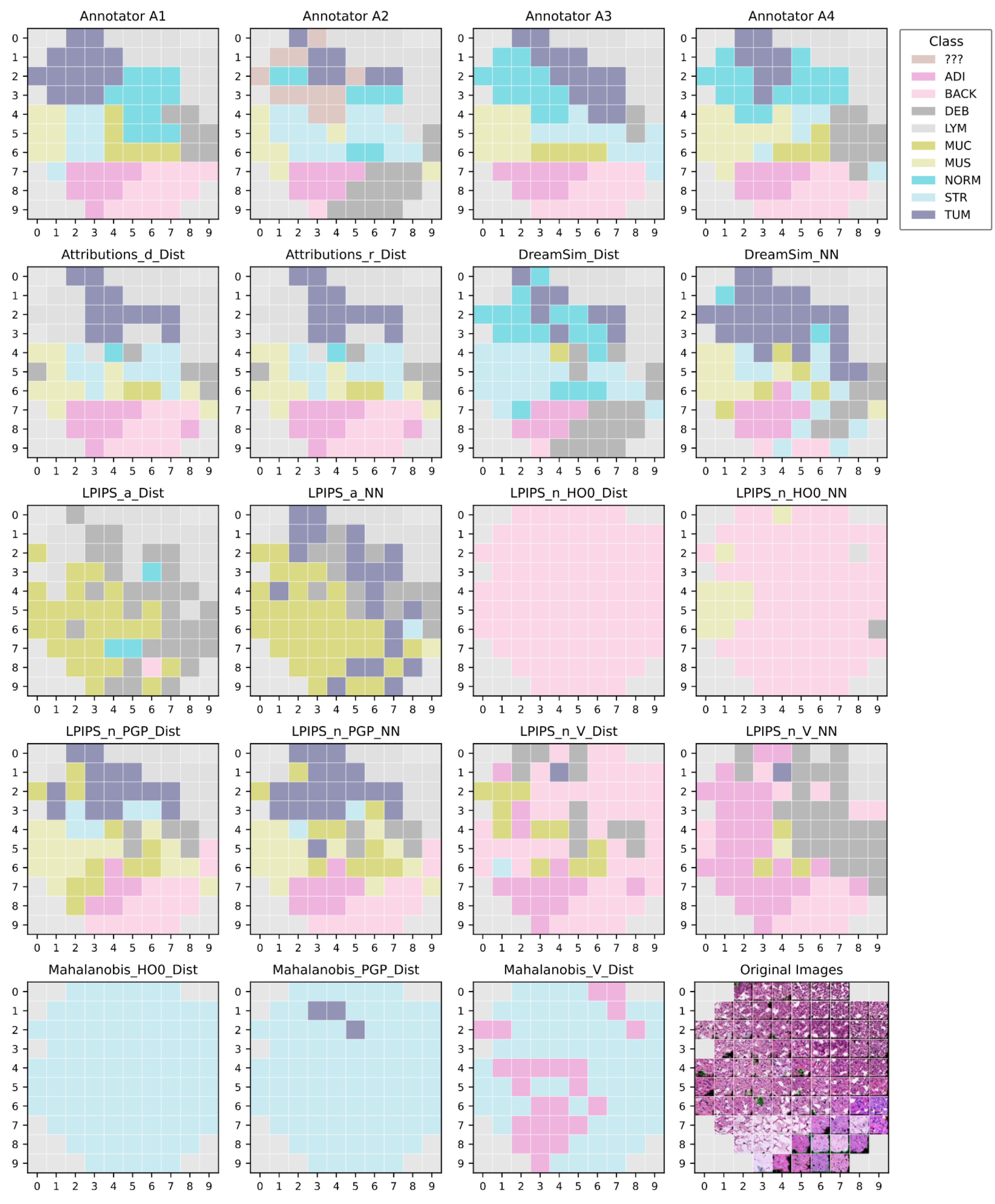

**Fig. 6 | Label maps for activation atlas cells in the National Center for Tumor Diseases tissue dataset (NCT).**

Cell-wise label maps for an activation atlas derived from the NCT dataset. Each grid shows the same activation atlas layout, in which each cell corresponds to an aggregated region in the reduced activation embedding space. Cell colors indicate the class label assigned to each activation atlas cell.
Top row: Independent annotations by four pathologists (Annotators A1–A4), assigning one of the nine NCT tissue classes: adipose tissue (ADI), background (BACK), debris (DEB), lymphocytes (LYM), mucus (MUC), smooth muscle (MUS), normal colon mucosa (NORM), cancer-associated stroma (STR), and colorectal adenocarcinoma epithelium (TUM), or an explicit label for uncertainty ("???").
Middle rows: Corresponding atlas label maps obtained using quantitative surrogate assignment methods. Attribution-based methods assign labels based on class attributions aggregated over atlas cells, using either distance-based aggregation (_Dist; assignment based on minimal attribution distance) or nearest-neighbor-based aggregation (_NN; assignment based on the closest individual activations). Perceptual similarity-based methods assign each atlas cell to the closest class using either DreamSim or Learned Perceptual Image Patch Similarity (LPIPS), again using distance-based (_Dist) or nearest-neighbor (_NN) strategies. LPIPS variants differ by the feature extractor used, indicated by suffixes _a (AlexNet), _H00 (H-optimus-0), _PGP (Prov-GigaPath), and _V (vision transformer, UNI). Mahalanobis distance-based methods assign labels based on class-conditional feature distributions computed in the corresponding representation spaces.
Bottom right: Generated image patches associated with each activation atlas cell, illustrating the underlying morphological content contributing to the aggregated activations.

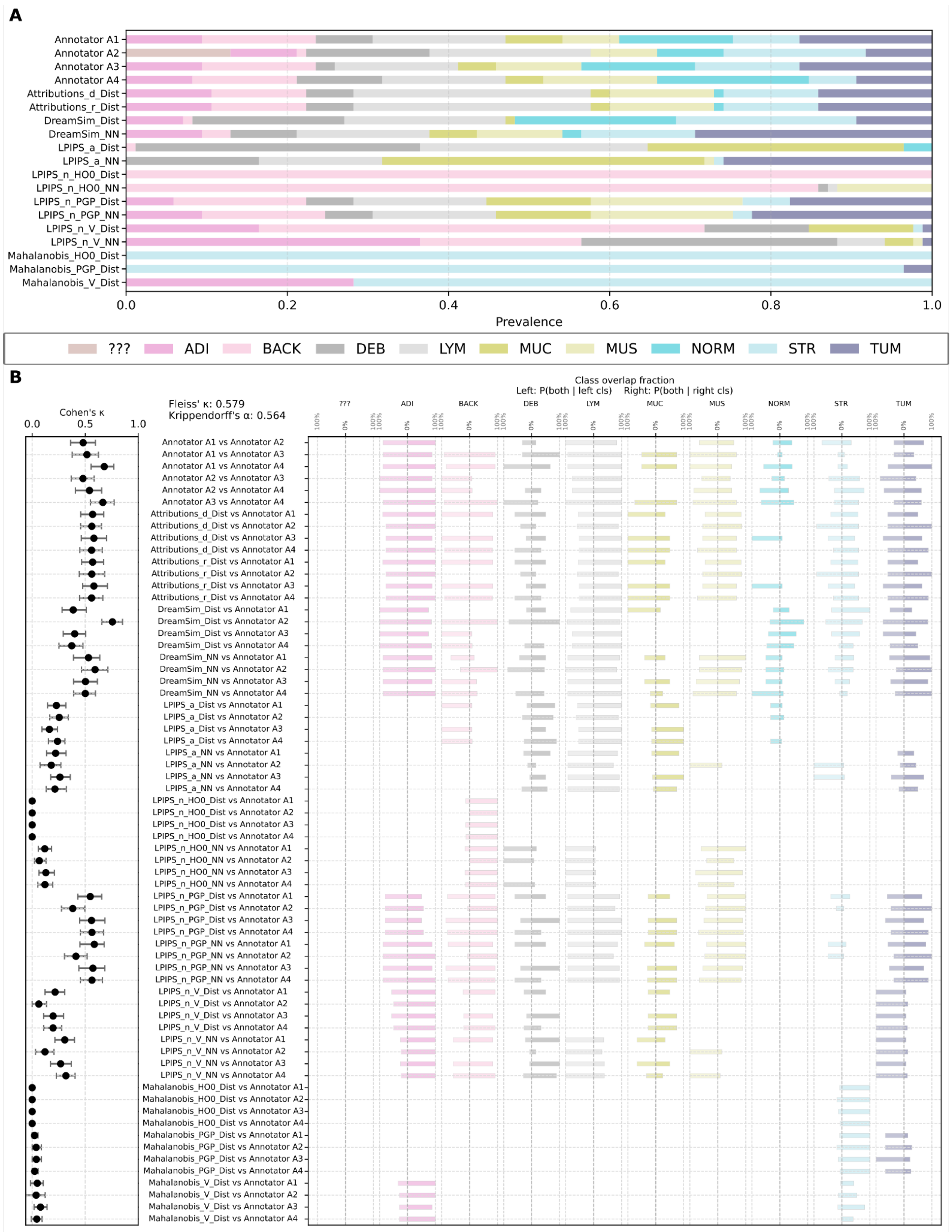

**Fig. 7 | Agreement and class coverage for activation atlas annotations in the National Center for Tumor Diseases (NCT) tissue task.**

Agreement analysis and class coverage for activation atlas cell annotations derived from the NCT tissue classification task.

**A** Class coverage showing the distribution of class assignments across activation atlas cells for each pathologist (Annotators A1–A4) and for each metric-based label assignment method. Bars indicate the proportion of atlas cells assigned to each NCT tissue class: adipose tissue (ADI), background (BACK), debris (DEB), lymphocytes (LYM), mucus (MUC), smooth muscle (MUS), normal colon mucosa (NORM), cancer-associated stroma (STR), and colorectal adenocarcinoma epithelium (TUM).

**B** Agreement metrics and class-wise overlap. Pairwise agreement between annotators and between annotators and metric-based assignments, computed excluding uncertain labels ("???") to ensure comparability of categorical agreement. Left: Cohen's κ values summarizing pairwise agreement, with points indicating mean κ across atlas cells and horizontal error bars denoting uncertainty estimated by bootstrap resampling. Overall inter-annotator agreement is summarized by Fleiss' κ and Krippendorff's α (top). Right: Class-wise overlap fractions indicating, for each tissue class, the fraction of atlas cells that received the same label from both compared sources. Bars on the left show, for each tissue class, how often atlas cells with that reference label are assigned the same class by the comparator. Bars on the right show, for each assigned class, how often the assigned cells originate from the corresponding reference label. Together, these measures capture asymmetric agreement and class-specific ambiguity across expert annotations and quantitative surrogate metrics.

# Tables

**Tab. 1: Links to directories containing the data and code used for the study.**

| | Link |
|---|---|
| NCT Dataset | https://zenodo.org/records/1214456 |
| TCGA Dataset | https://zenodo.org/records/5889558 |
| Code | https://github.com/KatherLab/PathoActivationAtlas |

**Tab. 2: Tissue types for the National Center for Tumor Diseases (NCT) dataset[40] and cancer types for the The Cancer Genome Atlas (TCGA) dataset[42].**

| Dataset | No. | Name | Abbreviation |
|---|---|---|---|
| NCT | 1 | Adipose | ADI |
| | 2 | Background | BACK |
| | 3 | Debris | DEB |
| | 4 | Lymphocytes | LYM |
| | 5 | Mucus | MUC |
| | 6 | Smooth muscle | MUS |
| | 7 | Normal colon mucosa | NORM |
| | 8 | Cancer-associated stroma | STR |
| | 9 | Colorectal adenocarcinoma epithelium | TUM |
| | 10 | Indecisive | ??? |
| TCGA | 1 | Breast invasive carcinoma | BRCA |
| | 2 | Colon adenocarcinoma | COAD |
| | 3 | Kidney renal clear cell carcinoma | KIRC |
| | 4 | Kidney renal papillary cell carcinoma | KIRP |
| | 5 | Lung adenocarcinoma | LUAD |
| | 6 | Lung squamous cell carcinoma | LUSC |
| | 7 | Lymphoid neoplasm diffuse large B-cell lymphoma | DLBCL |
| | 8 | Prostate adenocarcinoma | PRAD |
| | 9 | Rectum adenocarcinoma | READ |
| | 10 | Sarcoma | SARC |
| | 11 | Skin cutaneous melanoma | SKCM |
| | 12 | Indecisive | ??? |

**Tab. 3: Conducted experiments and results evaluated with the Area Under the Receiver Operating Characteristic curve (AUROC).** For all experiments, the UNI foundation model was used with frozen weights, and only the appended linear classification layer was trained.

| Model | Dataset | AUROC | F1 | Accuracy |
|---|---|---|---|---|
| UNI & linear classifier | NCT | 1.00 | 0.92 | 0.92 |
| UNI & linear classifier | TCGA-5 | 0.99 | 0.83 | 0.82 |
| UNI & linear classifier | TCGA-8 | 0.99 | 0.85 | 0.85 |
| UNI & linear classifier | TCGA-11 | 0.99 | 0.84 | 0.86 |

# Supplementary Figures

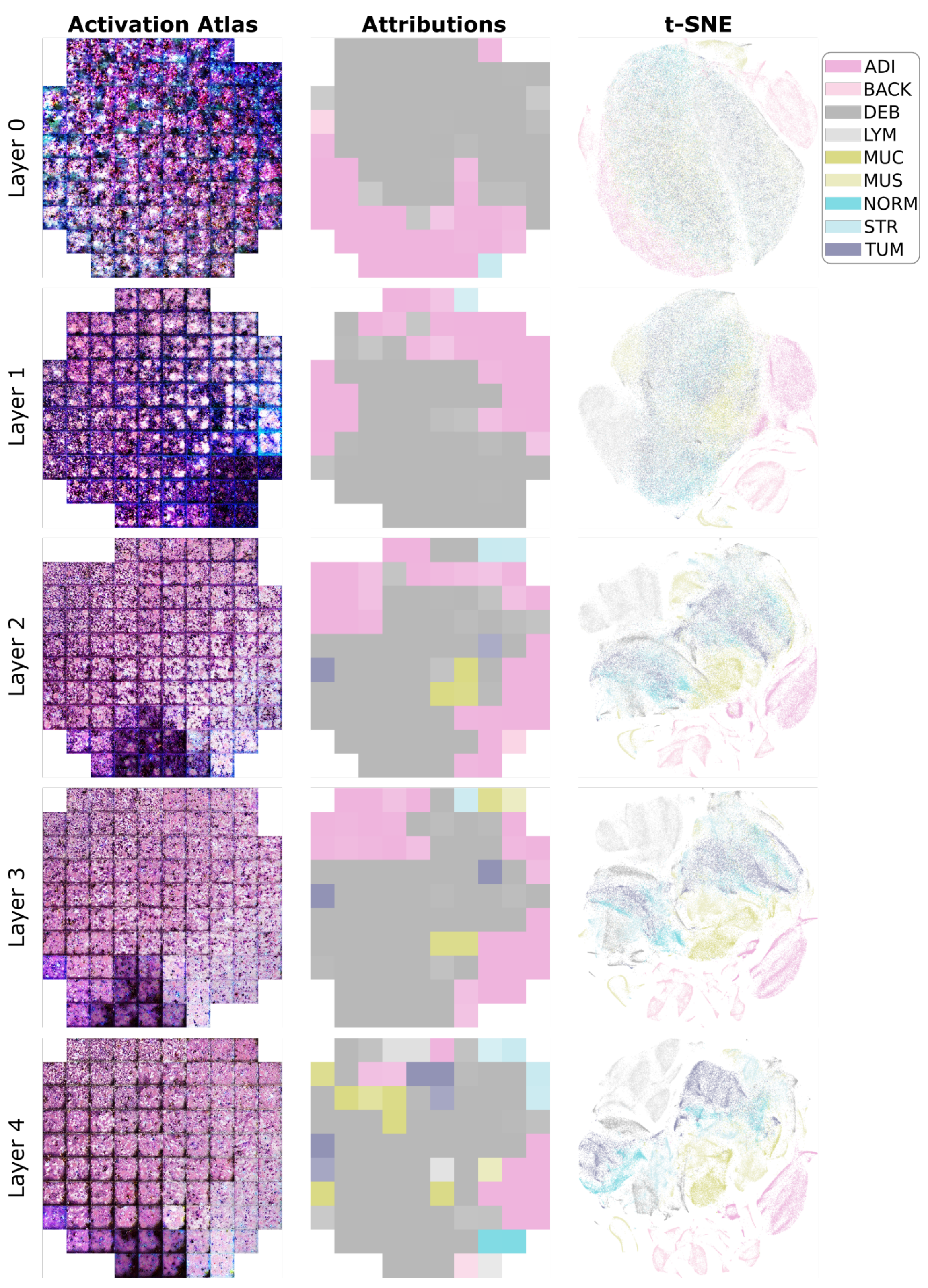

**Fig. S1 | Activation atlases, class attributions, and embedding projections for layers 0–4 in the National Center for Tumor Diseases (NCT) dataset.** Activation atlases, class attribution maps, and two-dimensional t-distributed stochastic neighbor embedding (t-SNE) projections derived from layers 0–4 (of a total of 24 layers) of the transformer-based UNI model trained on the NCT dataset. Each row corresponds to one network layer. The first column shows activation atlases generated by feature inversion, summarizing aggregated activation patterns across the embedding space. The second column shows tissue-type attribution maps indicating the dominant tissue class associated with each activation atlas cell. The third column shows t-SNE projections of patch-level feature embeddings extracted from the same layer, colored by tissue class. Class color codes are shown on the right, assigning one of the nine tissue classes: adipose tissue (ADI), background (BACK), debris (DEB), lymphocytes (LYM), mucus (MUC), smooth muscle (MUS), normal colon mucosa (NORM), cancer-associated stroma (STR), and colorectal adenocarcinoma epithelium (TUM).

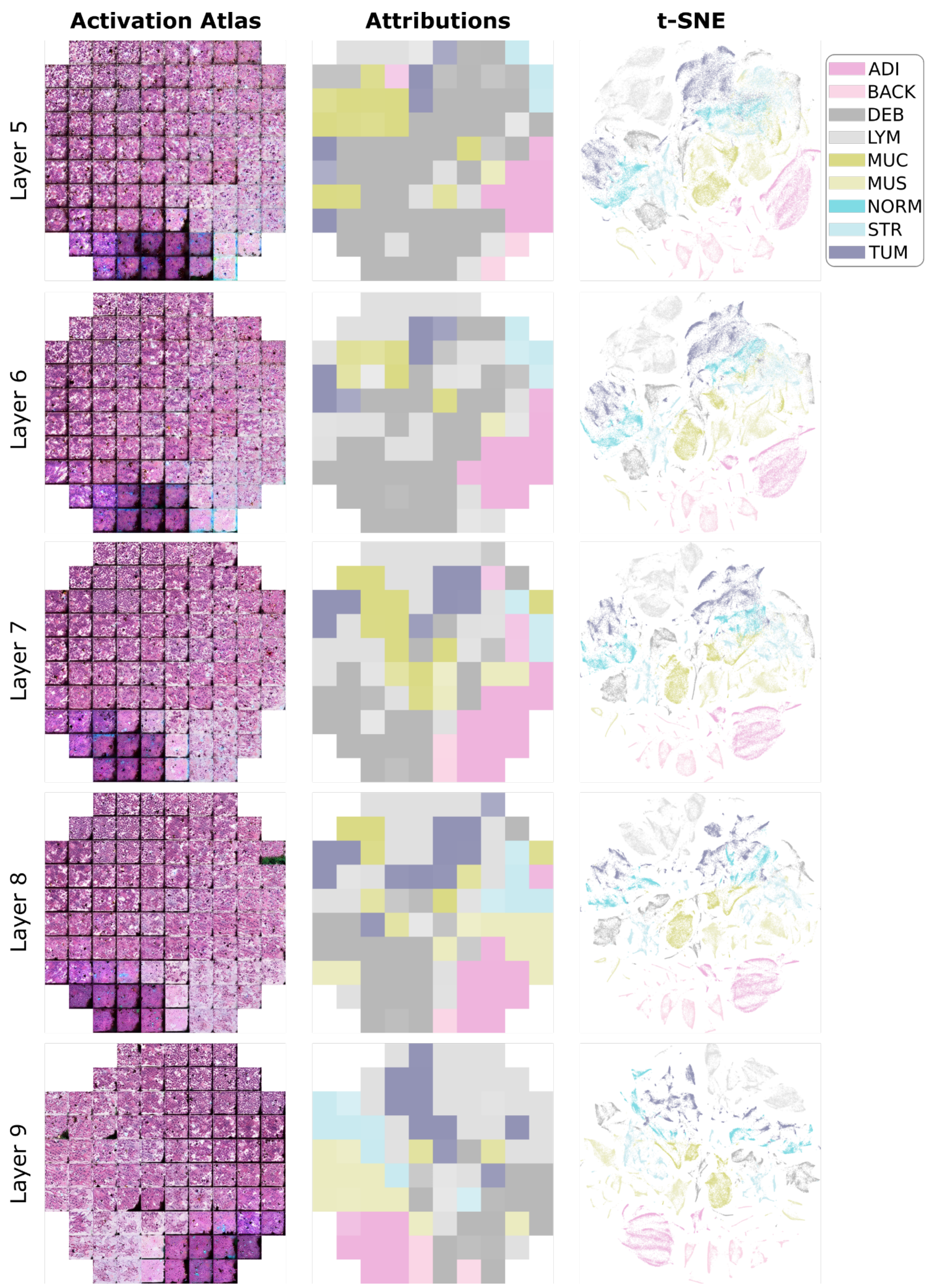

**Fig. S2 | Activation atlases, class attributions, and embedding projections for layers 5–9 in the National Center for Tumor Diseases (NCT) dataset.** Activation atlases, class attribution

maps, and two-dimensional t-distributed stochastic neighbor embedding (t-SNE) projections derived from layers 5–9 (of a total of 24 layers) of the transformer-based UNI model trained on the NCT dataset. Each row corresponds to one network layer. The first column shows activation atlases generated by feature inversion, summarizing aggregated activation patterns across the embedding space. The second column shows tissue-type attribution maps indicating the dominant tissue class associated with each activation atlas cell. The third column shows t-SNE projections of patch-level feature embeddings extracted from the same layer, colored by tissue class. Class color codes are shown on the right, assigning one of the nine tissue classes: adipose tissue (ADI), background (BACK), debris (DEB), lymphocytes (LYM), mucus (MUC), smooth muscle (MUS), normal colon mucosa (NORM), cancer-associated stroma (STR), and colorectal adenocarcinoma epithelium (TUM).

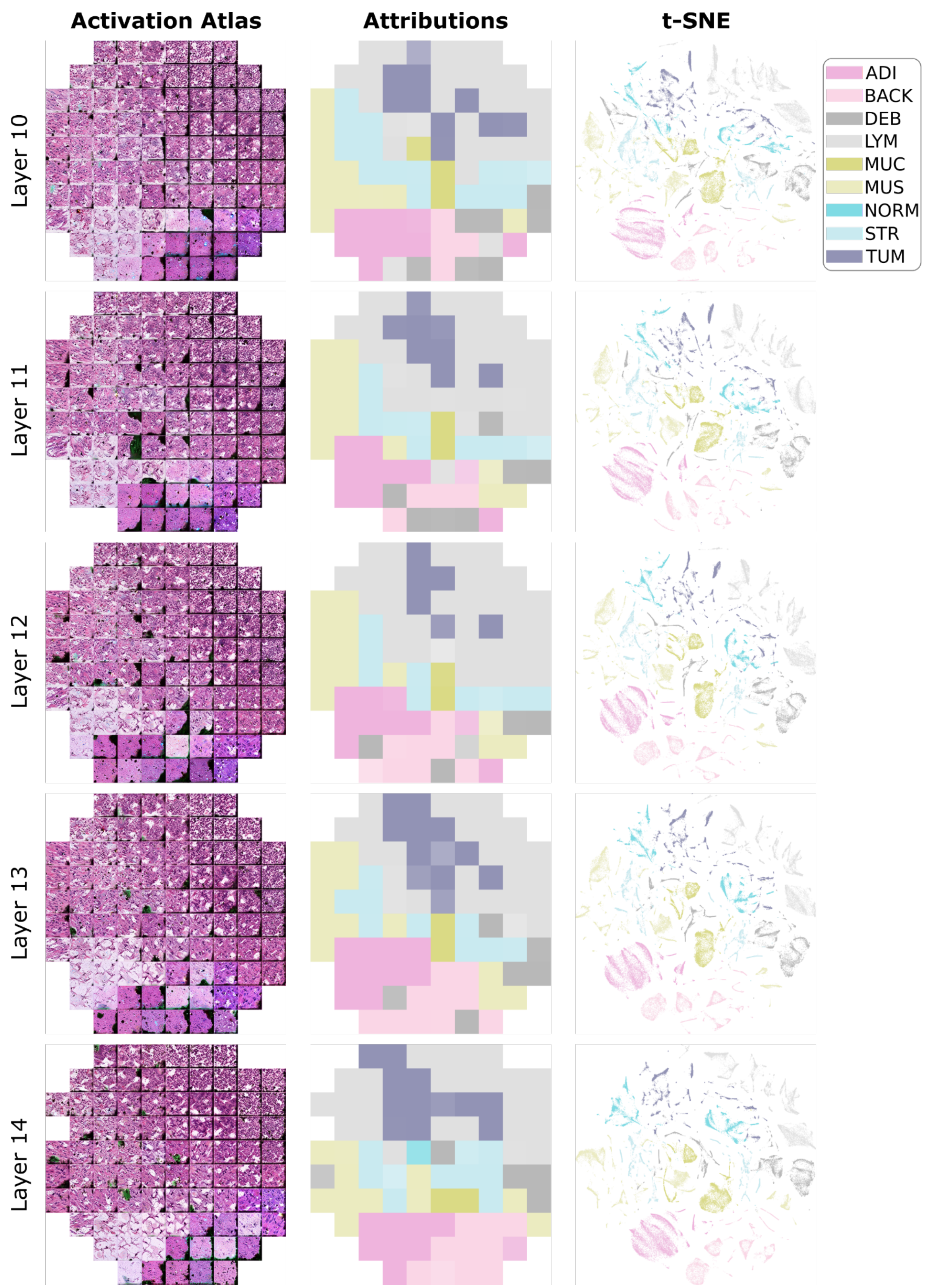

**Fig. S3 | Activation atlases, class attributions, and embedding projections for layers 10–14 in the National Center for Tumor Diseases (NCT) dataset.** Activation atlases, class attribution

maps, and two-dimensional t-distributed stochastic neighbor embedding (t-SNE) projections derived from layers 10–14 (of a total of 24 layers) of the transformer-based UNI model trained on the NCT dataset. Each row corresponds to one network layer. The first column shows activation atlases generated by feature inversion, summarizing aggregated activation patterns across the embedding space. The second column shows tissue-type attribution maps indicating the dominant tissue class associated with each activation atlas cell. The third column shows t-SNE projections of patch-level feature embeddings extracted from the same layer, colored by tissue class. Class color codes are shown on the right, assigning one of the nine tissue classes: adipose tissue (ADI), background (BACK), debris (DEB), lymphocytes (LYM), mucus (MUC), smooth muscle (MUS), normal colon mucosa (NORM), cancer-associated stroma (STR), and colorectal adenocarcinoma epithelium (TUM).

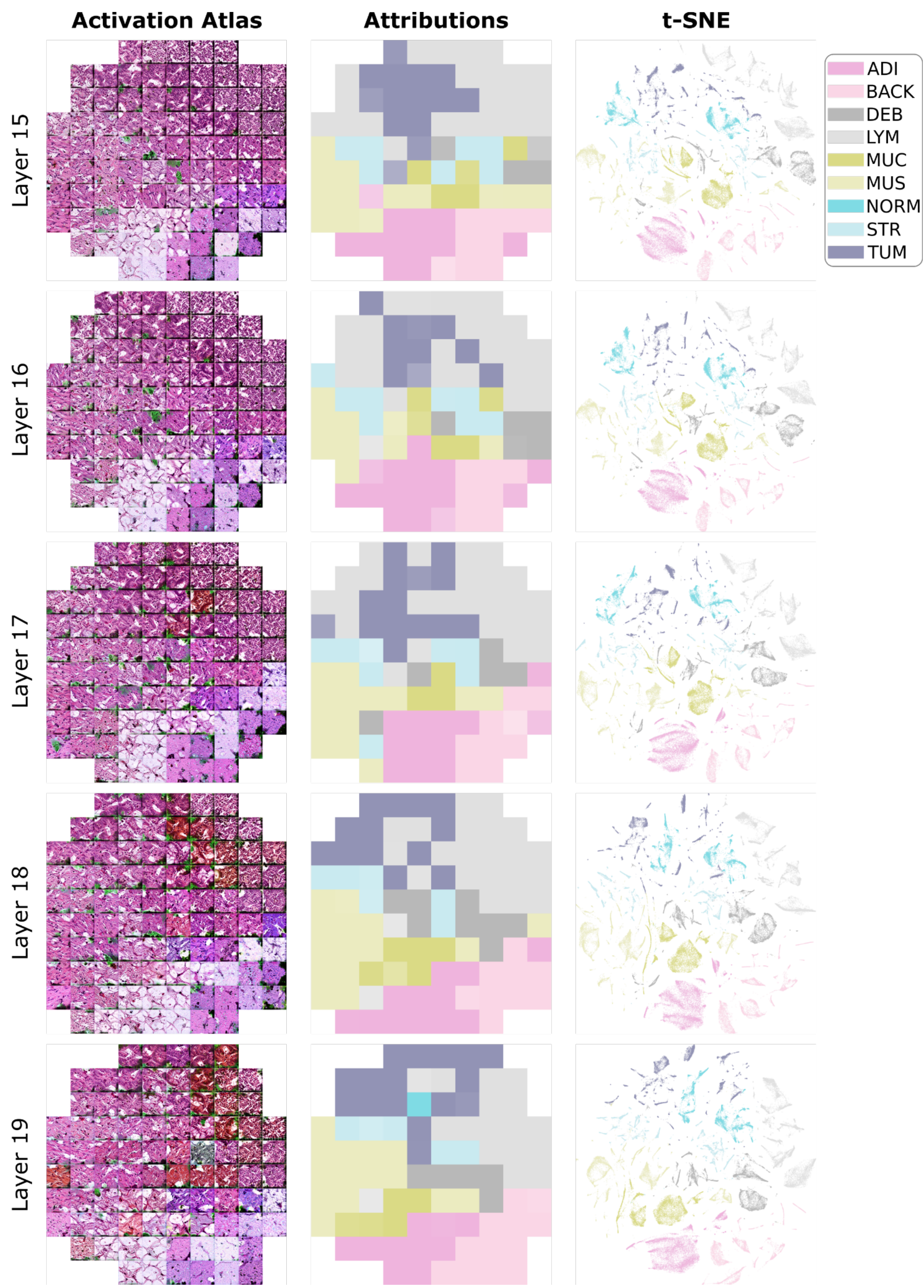

**Fig. S4 | Activation atlases, class attributions, and embedding projections for layers 15–19 in the National Center for Tumor Diseases (NCT) dataset.** Activation atlases, class attribution

maps, and two-dimensional t-distributed stochastic neighbor embedding (t-SNE) projections derived from layers 15–19 (of a total of 24 layers) of the transformer-based UNI model trained on the NCT dataset. Each row corresponds to one network layer. The first column shows activation atlases generated by feature inversion, summarizing aggregated activation patterns across the embedding space. The second column shows tissue-type attribution maps indicating the dominant tissue class associated with each activation atlas cell. The third column shows t-SNE projections of patch-level feature embeddings extracted from the same layer, colored by tissue class. Class color codes are shown on the right, assigning one of the nine tissue classes: adipose tissue (ADI), background (BACK), debris (DEB), lymphocytes (LYM), mucus (MUC), smooth muscle (MUS), normal colon mucosa (NORM), cancer-associated stroma (STR), and colorectal adenocarcinoma epithelium (TUM).

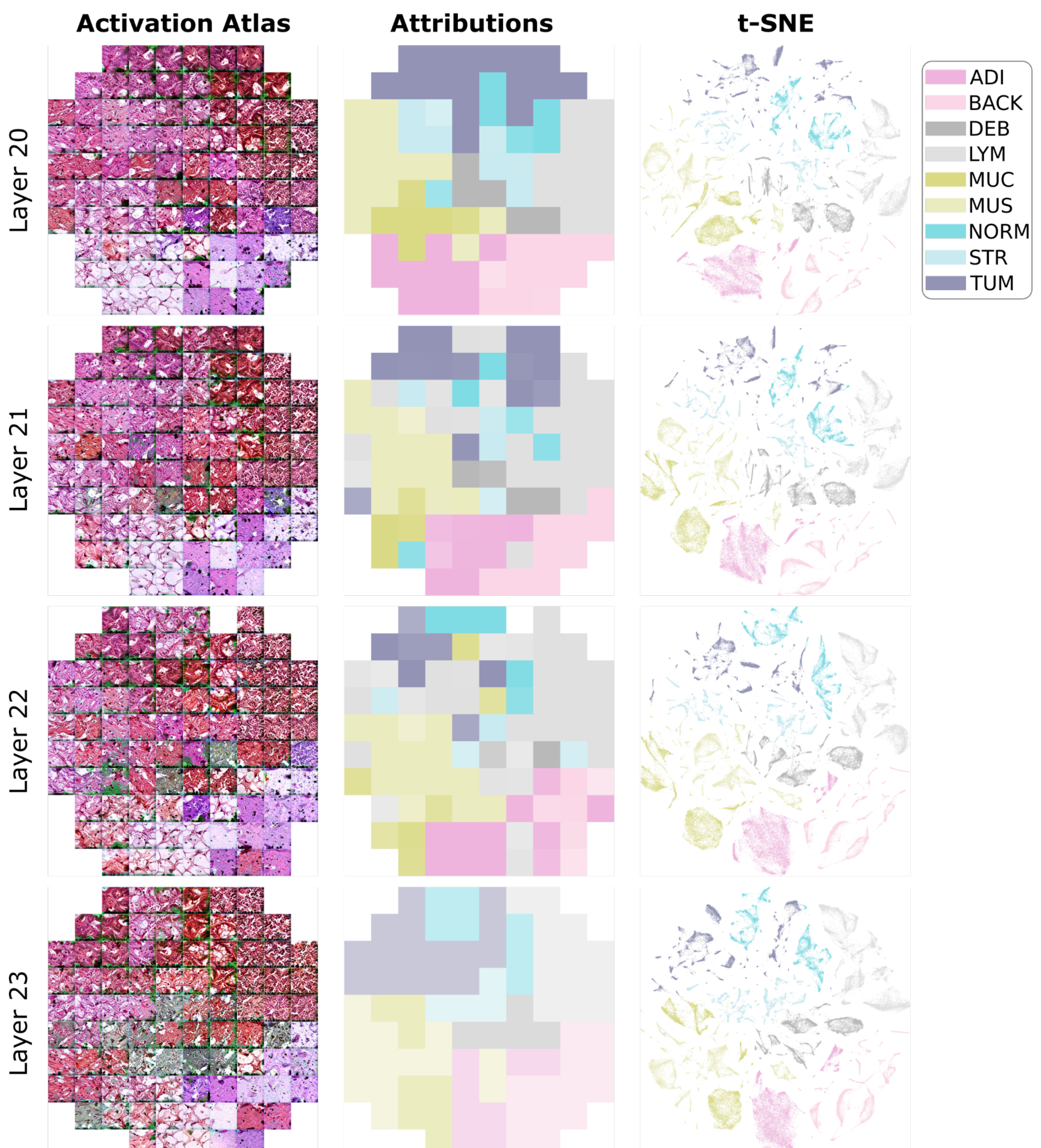

**Fig. S5 | Activation atlases, class attributions, and embedding projections for layers 20–23 in the National Center for Tumor Diseases (NCT) dataset.** Activation atlases, class attribution maps, and two-dimensional t-distributed stochastic neighbor embedding (t-SNE) projections derived from layers 20–23 (of a total of 24 layers) of the transformer-based UNI model trained on the NCT dataset. Each row corresponds to one network layer. The first column shows activation atlases generated by feature inversion, summarizing aggregated activation patterns across the embedding space. The second column shows tissue-type attribution maps indicating the dominant tissue class associated with each activation atlas cell. The third column shows t-SNE projections of patch-level feature embeddings extracted from the same layer, colored by tissue class. Class color codes are shown on the right, assigning one of the nine tissue classes: adipose tissue (ADI), background (BACK), debris (DEB), lymphocytes (LYM), mucus (MUC), smooth muscle

(MUS), normal colon mucosa (NORM), cancer-associated stroma (STR), and colorectal adeno-carcinoma epithelium (TUM).

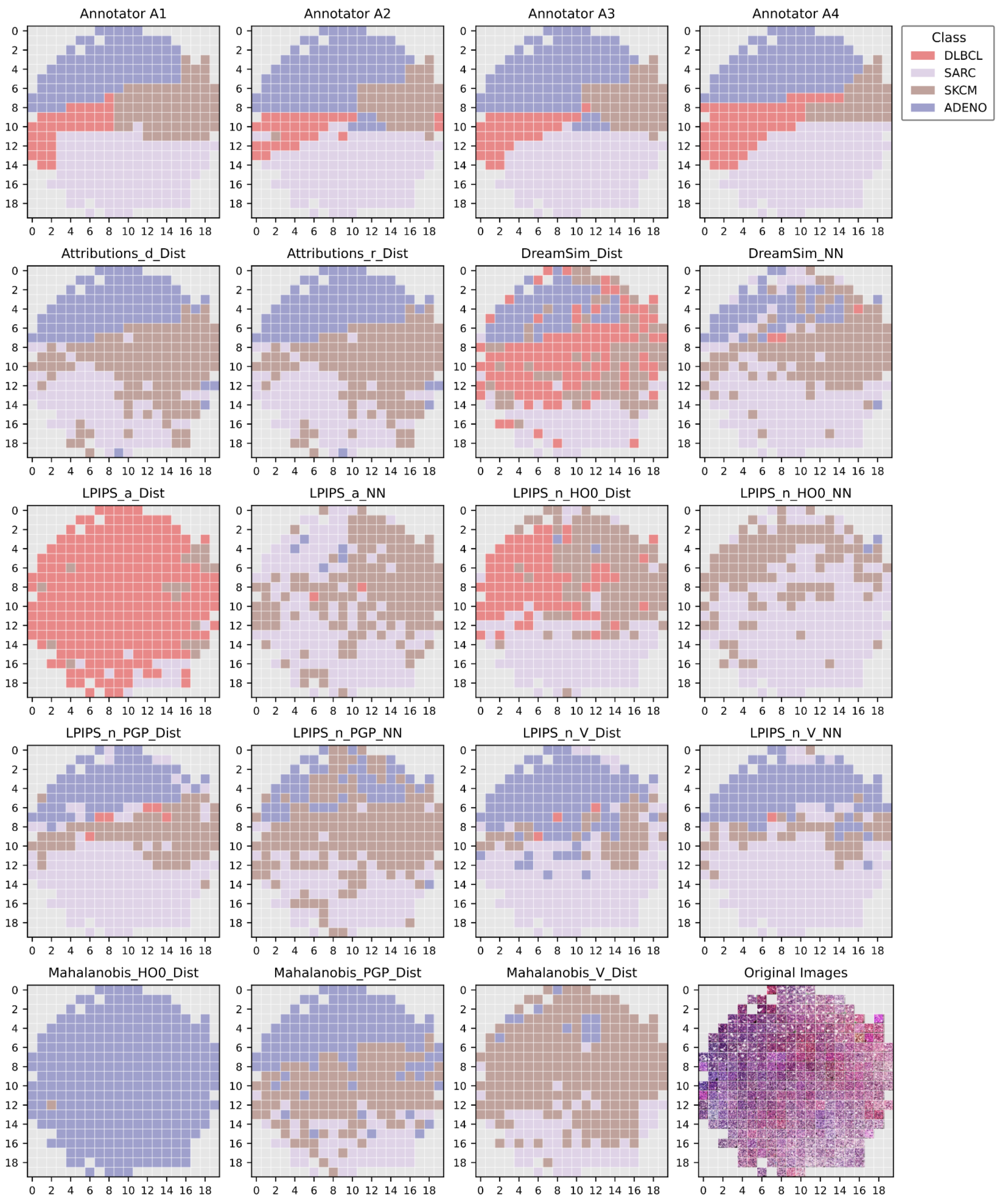

**Fig. S6 | Label maps for activation atlas cells in The Cancer Genome Atlas dataset (TCGA-5-adeno).**

Cell-wise label maps for an activation atlas derived from the TCGA adenocarcinoma-level classification task (TCGA-5-adeno). Each grid shows the same activation atlas layout, in which each cell corresponds to an aggregated region in the reduced activation embedding space. Cell colors indicate the class label assigned to each activation atlas cell.
Top row: Independent annotations by four pathologists (Annotators A1–A4), assigning one of the five TCGA cancer classes: adenocarcinoma (ADENO), lymphoid neoplasm diffuse large B-cell lymphoma (DLBCL), sarcoma (SARC), skin cutaneous melanoma (SKCM), and lung squamous cell carcinoma (LUSC).
Middle rows: Corresponding atlas label maps obtained using quantitative surrogate assignment methods. Attribution-based methods assign labels based on class attributions aggregated over atlas cells, using either distance-based aggregation (_Dist; assignment based on minimal attribution distance) or nearest-neighbor-based aggregation (_NN; assignment based on the closest individual activations). Perceptual similarity-based methods assign each atlas cell to the closest class using either DreamSim or Learned Perceptual Image Patch Similarity (LPIPS), again using distance-based (_Dist) or nearest-neighbor (_NN) strategies. LPIPS variants differ by the feature extractor used, indicated by suffixes _a (AlexNet), _H00 (H-optimus-0), _PGP (Prov-GigaPath), and _V (vision transformer, UNI). Mahalanobis distance-based methods assign labels based on class-conditional feature distributions computed in the corresponding representation spaces.
Bottom right: Generated image patches associated with each activation atlas cell, illustrating the underlying morphological content contributing to the aggregated activations.

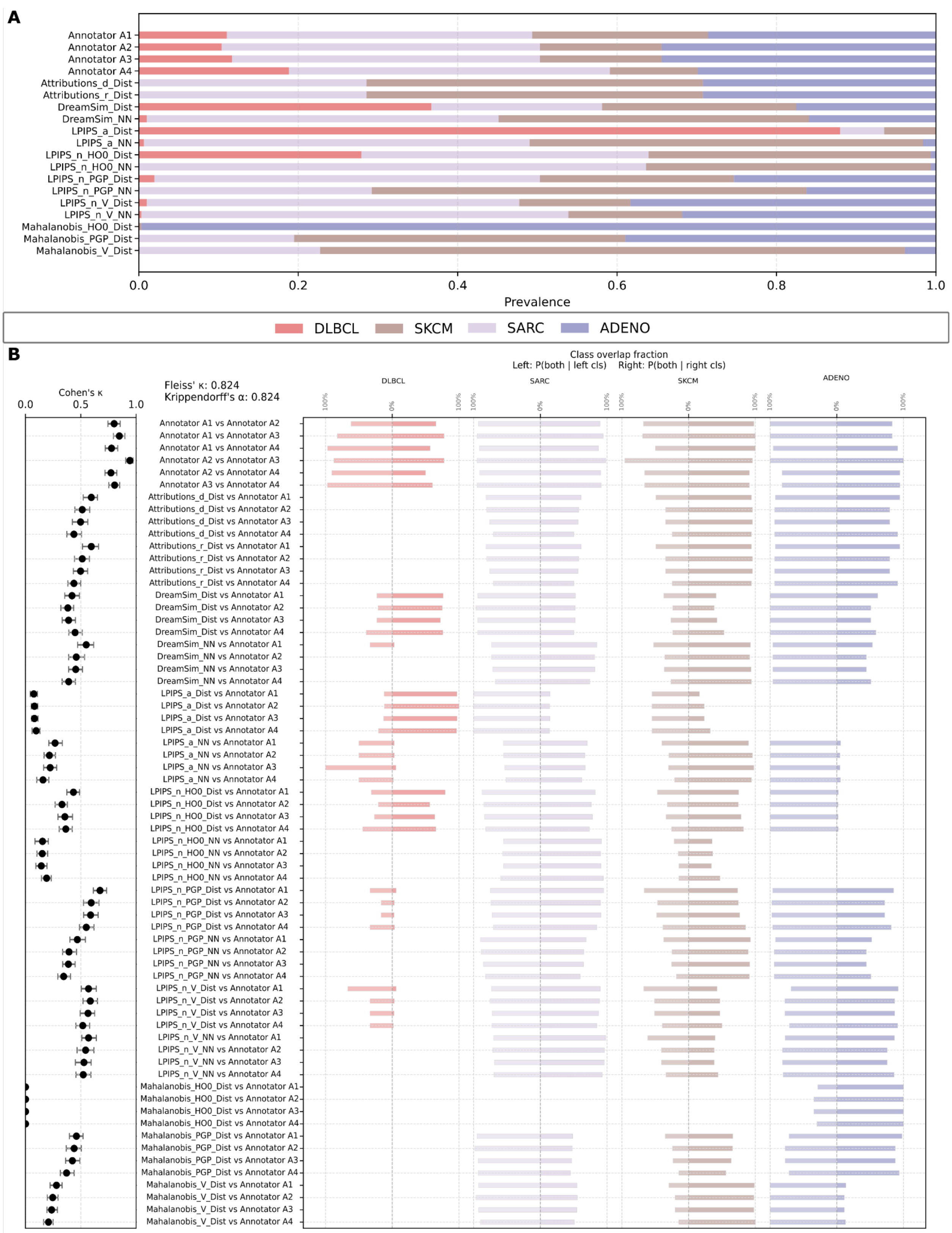

**Fig. S7 | Agreement and class coverage for activation atlas annotations in The Cancer Genome Atlas (TCGA-5-adeno) cancer task.**

Agreement analysis and class coverage for activation atlas cell annotations derived from the TCGA-5 adenocarcinoma-level classification task.

**A**. Class coverage. Stacked bar plots showing the distribution of class assignments across activation atlas cells for each pathologist (Annotators A1–A4) and for each metric-based label assignment method. Bars indicate the proportion of atlas cells assigned to each TCGA-5 cancer class: adenocarcinoma (ADENO), lymphoid neoplasm diffuse large B-cell lymphoma (DLBCL), sarcoma (SARC), and skin cutaneous melanoma (SKCM).

**B**. Agreement metrics and class-wise overlap. Pairwise agreement between annotators and between annotators and metric-based assignments, computed excluding uncertain labels to ensure comparability of categorical agreement. Left: Cohen's $\kappa$ values summarizing pairwise agreement, with points indicating mean $\kappa$ across activation atlas cells and horizontal error bars denoting uncertainty estimated by bootstrap resampling. Overall inter-annotator agreement is summarized by Fleiss' $\kappa$ and Krippendorff's $\alpha$ (top). Right: Class-wise overlap fractions indicating, for each cancer class, the fraction of activation atlas cells that received the same label from both compared sources. Bars on the left show, for each reference class, how often atlas cells with that reference label are assigned the same class by the comparator. Bars on the right show, for each assigned class, how often the assigned atlas cells originate from the corresponding reference class. Together, these measures capture asymmetric agreement and class-specific ambiguity across expert annotations and quantitative surrogate metrics.

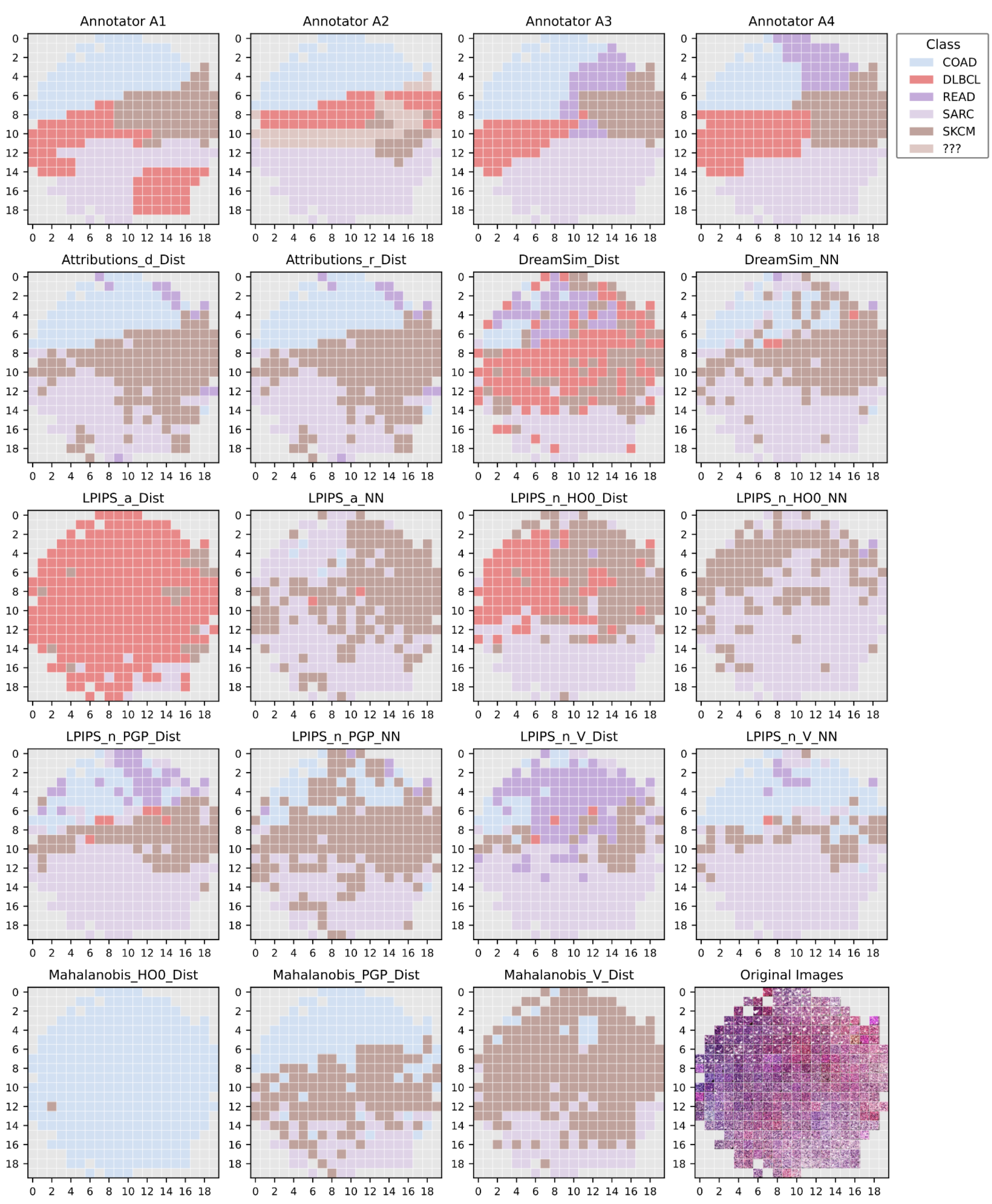

**Fig. S8 | Label maps for activation atlas cells in The Cancer Genome Atlas dataset (TCGA-5-subclass).**

Cell-wise label maps for an activation atlas derived from the TCGA-5 subclass-level classification task. Each grid shows the same activation atlas layout, in which each cell corresponds to an aggregated region in the reduced activation embedding space. Cell colors indicate the class label assigned to each activation atlas cell.
Top row: Independent annotations by four pathologists (Annotators A1–A4), assigning one of the five TCGA cancer subclasses: colon adenocarcinoma (COAD), lymphoid neoplasm diffuse large B-cell lymphoma (DLBCL), rectum adenocarcinoma (READ), sarcoma (SARC), and skin cutaneous melanoma (SKCM), or an explicit label for uncertainty ("???").
Middle rows: Corresponding atlas label maps obtained using quantitative surrogate assignment methods. Attribution-based methods assign labels based on class attributions aggregated over atlas cells, using either distance-based aggregation (_Dist; assignment based on minimal attribution distance) or nearest-neighbor-based aggregation (_NN; assignment based on the closest individual activations). Perceptual similarity-based methods assign each atlas cell to the closest class using either DreamSim or Learned Perceptual Image Patch Similarity (LPIPS), again using distance-based (_Dist) or nearest-neighbor (_NN) strategies. LPIPS variants differ by the feature extractor used, indicated by suffixes _a (AlexNet), _H00 (H-optimus-0), _PGP (Prov-GigaPath), and _V (vision transformer, UNI). Mahalanobis distance-based methods assign labels based on class-conditional feature distributions computed in the corresponding representation spaces.
Bottom right: Generated image patches associated with each activation atlas cell, illustrating the underlying morphological content contributing to the aggregated activations.

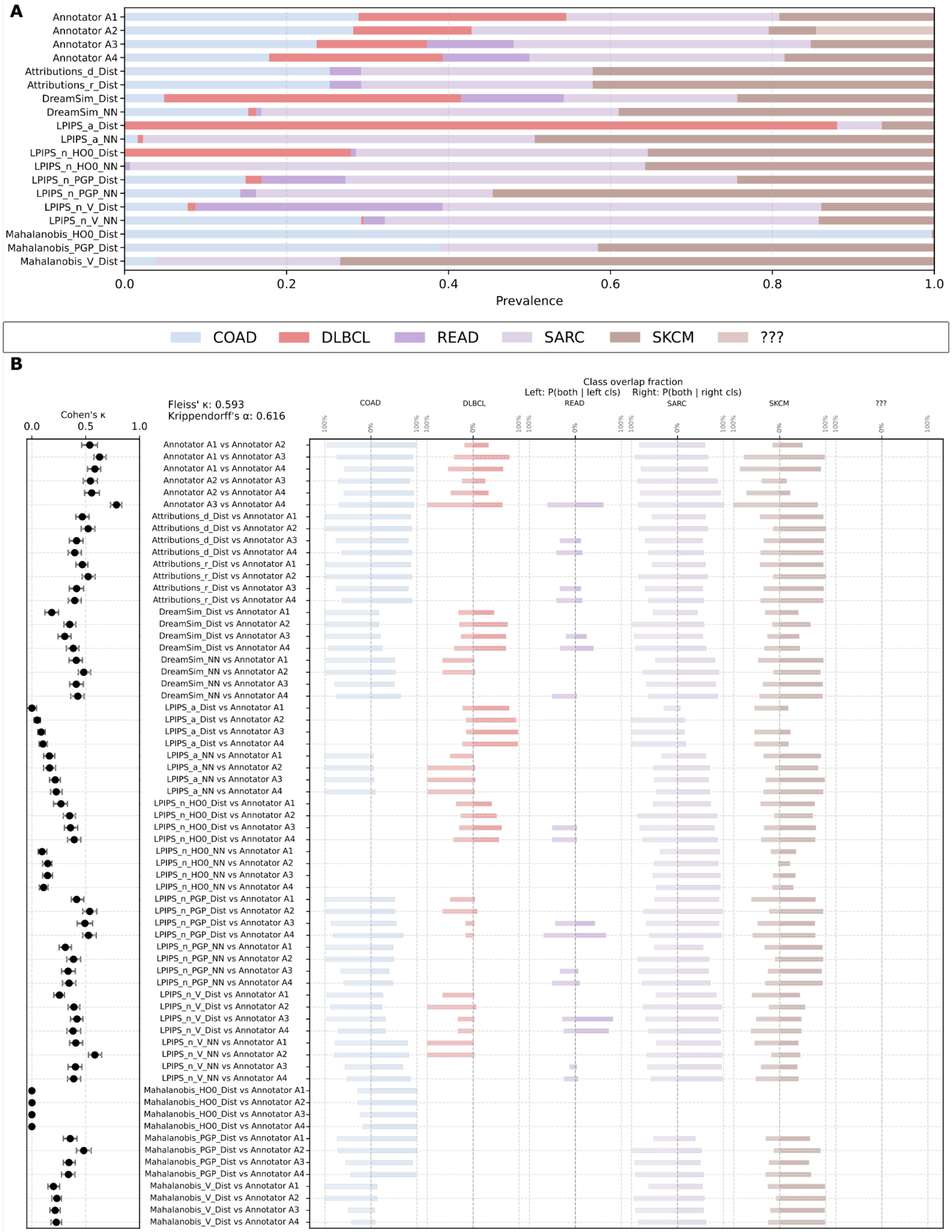

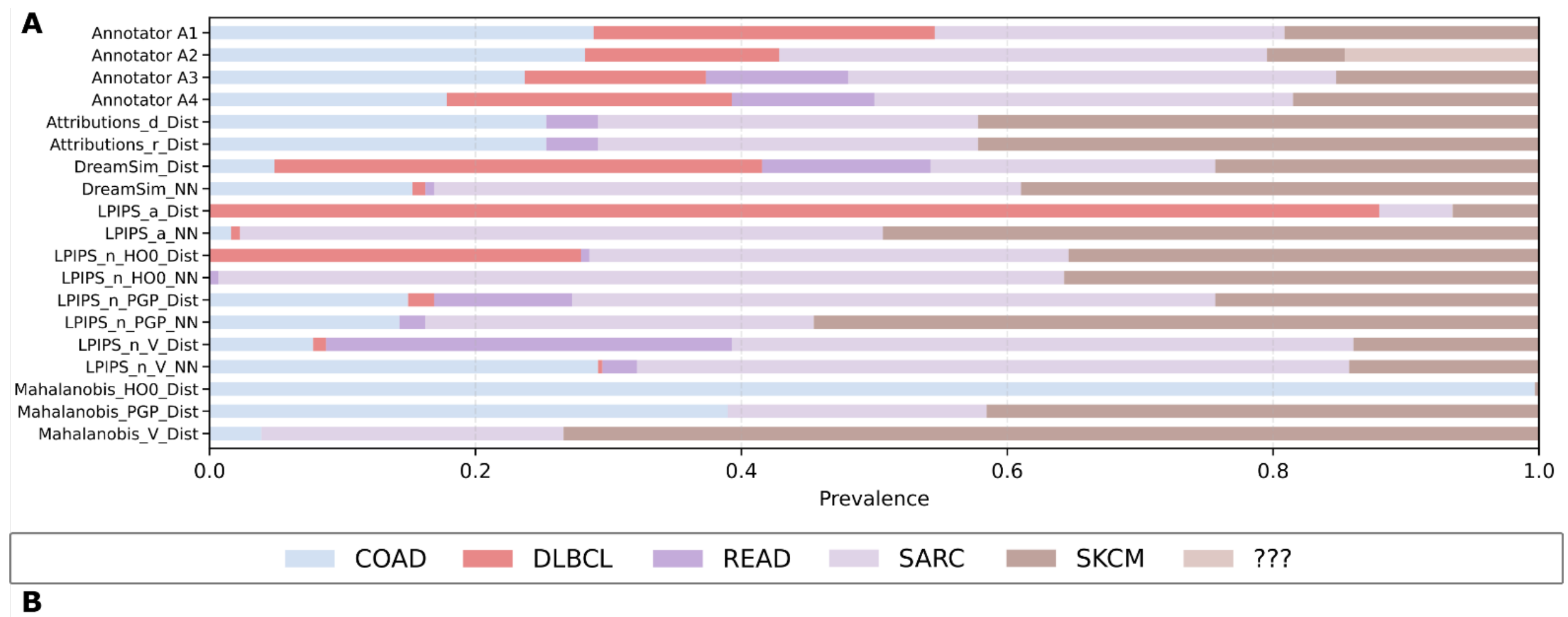

B

Cohen's κ
0.0 0.5 1.0

Fleiss' κ: 0.593
Krippendorff's α: 0.616

Class overlap fraction
Left: P(both | left cls) Right: P(both | right cls)

COAD DLBCL READ SARC SKCM ???

100% 0% 100% 100% 0% 100% 100% 0% 100% 100% 0% 100% 100% 0% 100% 100% 0% 100%

Annotator A1 vs Annotator A2
Annotator A1 vs Annotator A3
Annotator A1 vs Annotator A4
Annotator A2 vs Annotator A3
Annotator A2 vs Annotator A4
Annotator A3 vs Annotator A4
Attributions_d_Dist vs Annotator A1
Attributions_d_Dist vs Annotator A2
Attributions_d_Dist vs Annotator A3
Attributions_d_Dist vs Annotator A4
Attributions_r_Dist vs Annotator A1
Attributions_r_Dist vs Annotator A2
Attributions_r_Dist vs Annotator A3
Attributions_r_Dist vs Annotator A4
DreamSim_Dist vs Annotator A1
DreamSim_Dist vs Annotator A2
DreamSim_Dist vs Annotator A3
DreamSim_Dist vs Annotator A4
DreamSim_NN vs Annotator A1
DreamSim_NN vs Annotator A2
DreamSim_NN vs Annotator A3
DreamSim_NN vs Annotator A4
LPIPS_a_Dist vs Annotator A1
LPIPS_a_Dist vs Annotator A2
LPIPS_a_Dist vs Annotator A3
LPIPS_a_Dist vs Annotator A4
LPIPS_a_NN vs Annotator A1
LPIPS_a_NN vs Annotator A2
LPIPS_a_NN vs Annotator A3
LPIPS_a_NN vs Annotator A4
LPIPS_n_HO0_Dist vs Annotator A1
LPIPS_n_HO0_Dist vs Annotator A2
LPIPS_n_HO0_Dist vs Annotator A3
LPIPS_n_HO0_Dist vs Annotator A4
LPIPS_n_HO0_NN vs Annotator A1
LPIPS_n_HO0_NN vs Annotator A2
LPIPS_n_HO0_NN vs Annotator A3
LPIPS_n_HO0_NN vs Annotator A4
LPIPS_n_PGP_Dist vs Annotator A1
LPIPS_n_PGP_Dist vs Annotator A2
LPIPS_n_PGP_Dist vs Annotator A3
LPIPS_n_PGP_Dist vs Annotator A4
LPIPS_n_PGP_NN vs Annotator A1
LPIPS_n_PGP_NN vs Annotator A2
LPIPS_n_PGP_NN vs Annotator A3
LPIPS_n_PGP_NN vs Annotator A4
LPIPS_n_V_Dist vs Annotator A1
LPIPS_n_V_Dist vs Annotator A2
LPIPS_n_V_Dist vs Annotator A3
LPIPS_n_V_Dist vs Annotator A4
LPIPS_n_V_NN vs Annotator A1
LPIPS_n_V_NN vs Annotator A2
LPIPS_n_V_NN vs Annotator A3
LPIPS_n_V_NN vs Annotator A4
Mahalanobis_HO0_Dist vs Annotator A1
Mahalanobis_HO0_Dist vs Annotator A2
Mahalanobis_HO0_Dist vs Annotator A3
Mahalanobis_HO0_Dist vs Annotator A4
Mahalanobis_PGP_Dist vs Annotator A1
Mahalanobis_PGP_Dist vs Annotator A2
Mahalanobis_PGP_Dist vs Annotator A3
Mahalanobis_PGP_Dist vs Annotator A4
Mahalanobis_V_Dist vs Annotator A1
Mahalanobis_V_Dist vs Annotator A2
Mahalanobis_V_Dist vs Annotator A3
Mahalanobis_V_Dist vs Annotator A4

**Fig. S9 | Agreement and class coverage for activation atlas annotations in The Cancer Genome Atlas (TCGA-5 subclass) cancer task.**

Agreement analysis and class coverage for activation atlas cell annotations derived from the TCGA-5 subclass-level classification task.

**A**. Class coverage. Stacked bar plots showing the distribution of class assignments across activation atlas cells for each pathologist (Annotators A1–A4) and for each metric-based label assignment method. Bars indicate the proportion of atlas cells assigned to each TCGA-5 cancer subclass: colon adenocarcinoma (COAD), lymphoid neoplasm diffuse large B-cell lymphoma (DLBCL), rectum adenocarcinoma (READ), sarcoma (SARC), and skin cutaneous melanoma (SKCM). Where present, uncertain annotations ("???") are included as an explicit category to reflect overall class coverage.

**B**. Agreement metrics and class-wise overlap. Pairwise agreement between annotators and between annotators and metric-based assignments, computed excluding uncertain labels ("???") to ensure comparability of categorical agreement. Left: Cohen's κ values summarizing pairwise agreement, with points indicating mean κ across activation atlas cells and horizontal error bars denoting uncertainty estimated by bootstrap resampling. Overall inter-annotator agreement is summarized by Fleiss' κ and Krippendorff's α (top). Right: Class-wise overlap fractions indicating, for each cancer subclass, the fraction of activation atlas cells that received the same label from both compared sources. Bars on the left show, for each reference subclass, how often atlas cells with that reference label are assigned the same subclass by the comparator. Bars on the right show, for each assigned subclass, how often the assigned atlas cells originate from the corresponding reference subclass. Together, these measures capture asymmetric agreement and subclass-specific ambiguity across expert annotations and quantitative surrogate metrics.

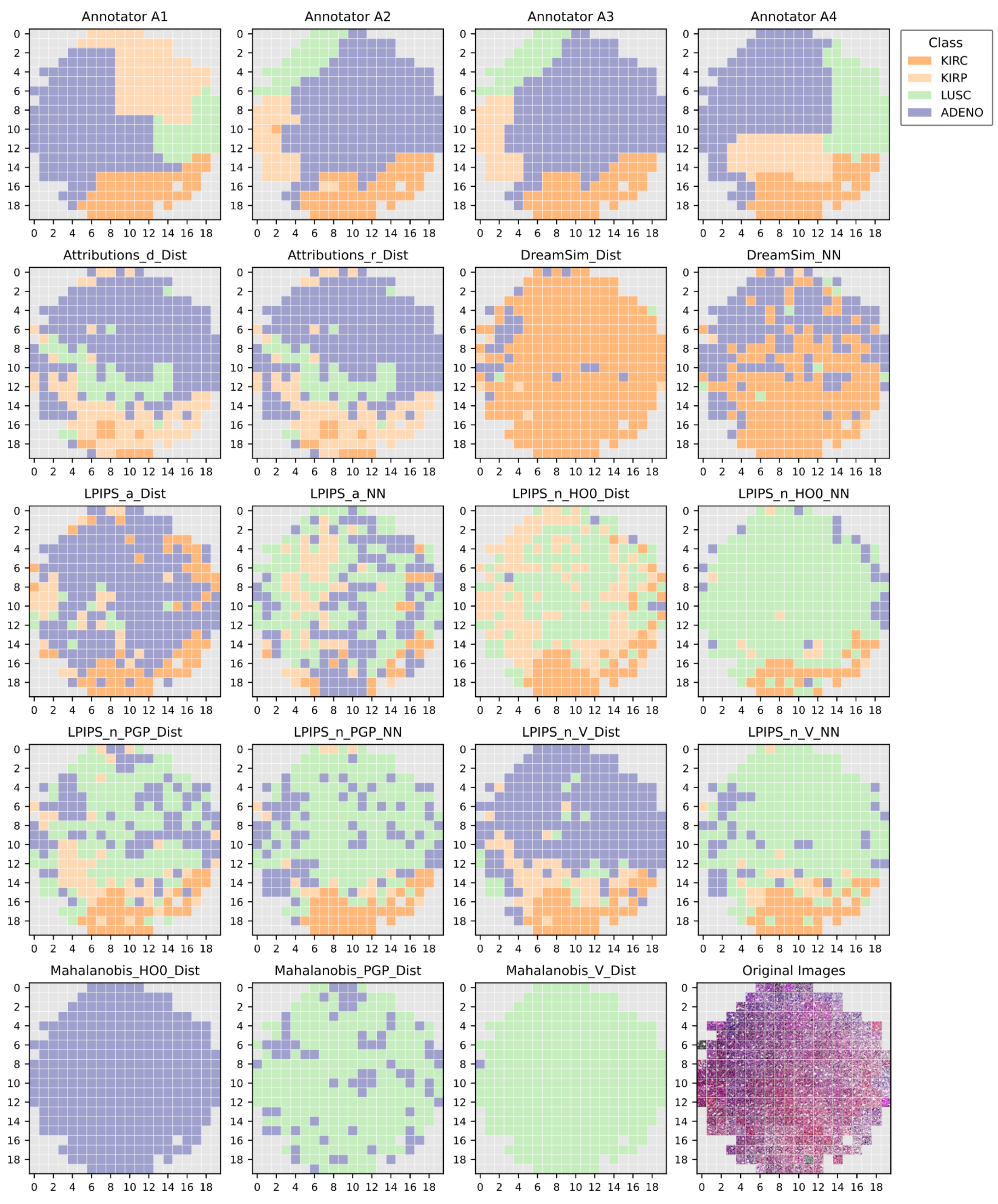

**Fig. S10 | Label maps for activation atlas cells in The Cancer Genome Atlas dataset (TCGA-8-adeno).**
Cell-wise label maps for an activation atlas derived from the TCGA-8 adenocarcinoma-level classification task. Each grid shows the same activation atlas layout, in which each cell corresponds to an aggregated region in the reduced activation embedding space. Cell colors indicate the class label assigned to each activation atlas cell.
Top row: Independent annotations by four pathologists (Annotators A1–A4), assigning one of the four TCGA cancer classes: adenocarcinoma (ADENO), kidney renal clear cell carcinoma (KIRC), kidney renal papillary cell carcinoma (KIRP), and lung squamous cell carcinoma (LUSC).
Middle rows: Corresponding atlas label maps obtained using quantitative surrogate assignment methods. Attribution-based methods assign labels based on class attributions aggregated over atlas cells, using either distance-based aggregation (_Dist; assignment based on minimal attribution distance) or nearest-neighbor-based aggregation (_NN; assignment based on the closest individual activations). Perceptual similarity-based methods assign each atlas cell to the closest class using either DreamSim or Learned Perceptual Image Patch Similarity (LPIPS), again using distance-based (_Dist) or nearest-neighbor (_NN) strategies. LPIPS variants differ by the feature extractor used, indicated by suffixes _a (AlexNet), _H00 (H-optimus-0), _PGP (Prov-GigaPath), and _V (vision transformer, UNI). Mahalanobis distance-based methods assign labels based on class-conditional feature distributions computed in the corresponding representation spaces.
Bottom right: Generated image patches associated with each activation atlas cell, illustrating the underlying morphological content contributing to the aggregated activations.

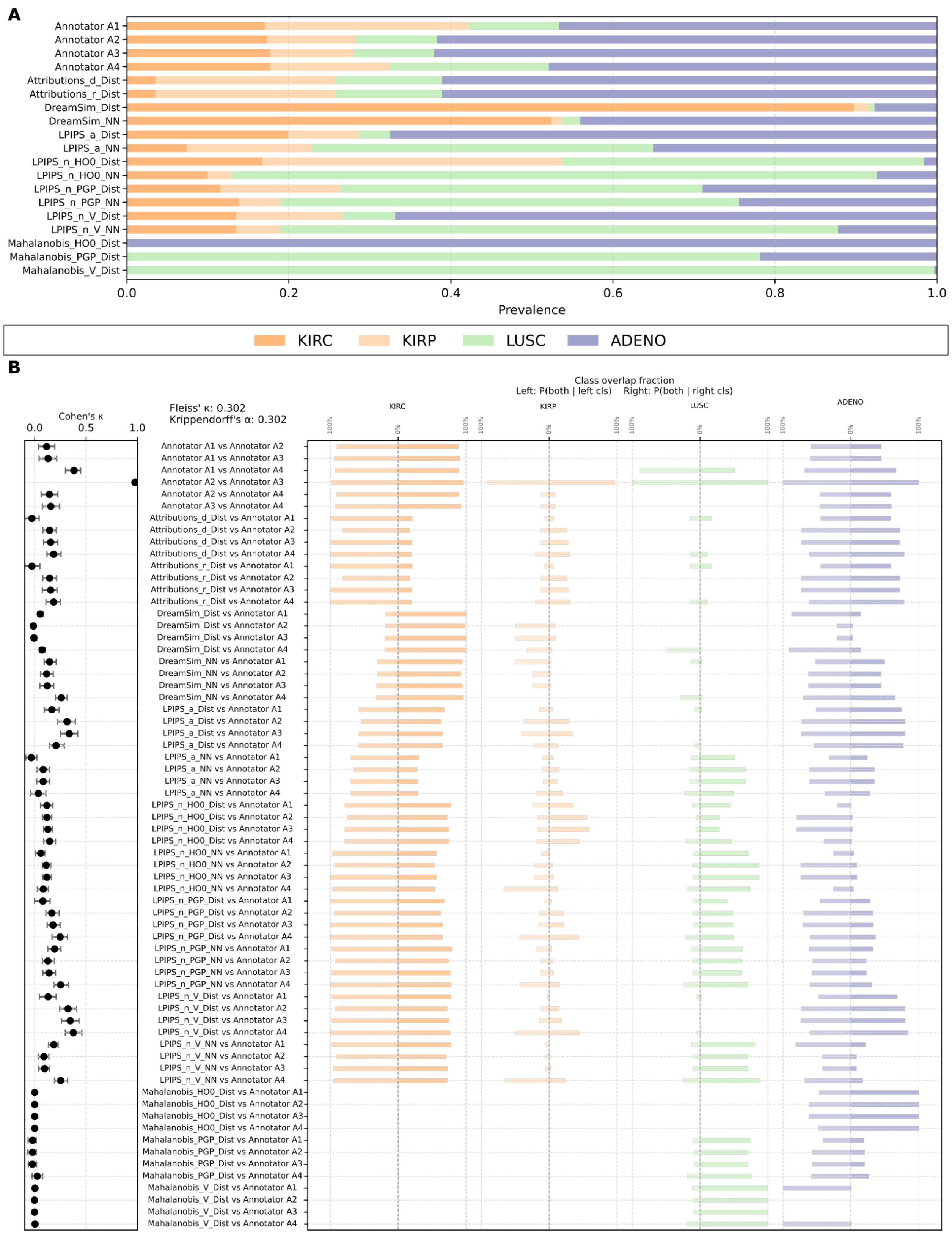

**Fig. S11 | Agreement and class coverage for activation atlas annotations in The Cancer Genome Atlas (TCGA-8 adeno) cancer task.**

Agreement analysis and class coverage for activation atlas cell annotations derived from the TCGA-8 adenocarcinoma-level classification task.

**A**. Class coverage. Stacked bar plots showing the distribution of class assignments across activation atlas cells for each pathologist (Annotators A1–A4) and for each metric-based label assignment method. Bars indicate the proportion of atlas cells assigned to each TCGA-8 cancer class: kidney renal clear cell carcinoma (KIRC), kidney renal papillary cell carcinoma (KIRP), lung squamous cell carcinoma (LUSC), and adenocarcinoma (ADENO).

**B**. Agreement metrics and class-wise overlap. Pairwise agreement between annotators and between annotators and metric-based assignments, computed excluding uncertain labels to ensure comparability of categorical agreement. Left: Cohen's $\kappa$ values summarizing pairwise agreement, with points indicating mean $\kappa$ across activation atlas cells and horizontal error bars denoting uncertainty estimated by bootstrap resampling. Overall inter-annotator agreement is summarized by Fleiss' $\kappa$ and Krippendorff's $\alpha$ (top). Right: Class-wise overlap fractions indicating, for each cancer class, the fraction of activation atlas cells that received the same label from both compared sources. Bars on the left show, for each reference class, how often atlas cells with that reference label are assigned the same class by the comparator. Bars on the right show, for each assigned class, how often the assigned atlas cells originate from the corresponding reference class. Together, these measures capture asymmetric agreement and class-specific ambiguity across expert annotations and quantitative surrogate metrics.

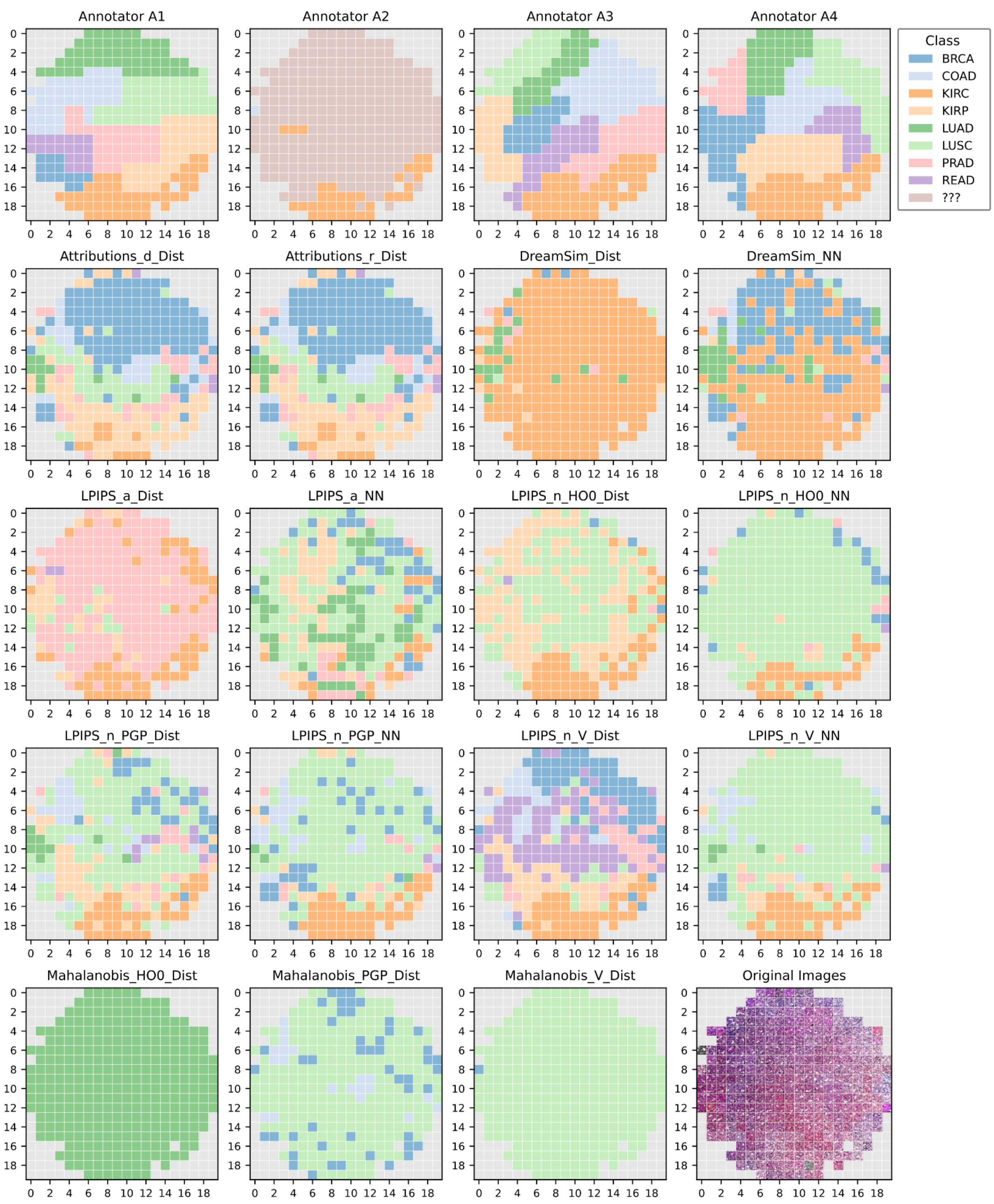

**Fig. S12 | Label maps for activation atlas cells in The Cancer Genome Atlas dataset (TCGA-8-subclass).**

Cell-wise label maps for an activation atlas derived from the TCGA-8 subclass-level classification task. Each grid shows the same activation atlas layout, in which each cell corresponds to an aggregated region in the reduced activation embedding space. Cell colors indicate the class label assigned to each activation atlas cell.
Top row: Independent annotations by four pathologists (Annotators A1–A4), assigning one of the eight TCGA cancer subclasses: breast invasive carcinoma (BRCA), colon adenocarcinoma (COAD), kidney renal clear cell carcinoma (KIRC), kidney renal papillary cell carcinoma (KIRP), lung adenocarcinoma (LUAD), lung squamous cell carcinoma (LUSC), prostate adenocarcinoma (PRAD), and rectum adenocarcinoma (READ), or an explicit label for uncertainty ("???").
Middle rows: Corresponding atlas label maps obtained using quantitative surrogate assignment methods. Attribution-based methods assign labels based on class attributions aggregated over atlas cells, using either distance-based aggregation (_Dist; assignment based on minimal attribution distance) or nearest-neighbor-based aggregation (_NN; assignment based on the closest individual activations). Perceptual similarity-based methods assign each atlas cell to the closest class using either DreamSim or Learned Perceptual Image Patch Similarity (LPIPS), again using distance-based (_Dist) or nearest-neighbor (_NN) strategies. LPIPS variants differ by the feature extractor used, indicated by suffixes _a (AlexNet), _H00 (H-optimus-0), _PGP (Prov-GigaPath), and _V (vision transformer, UNI). Mahalanobis distance-based methods assign labels based on class-conditional feature distributions computed in the corresponding representation spaces.
Bottom right: Generated image patches associated with each activation atlas cell, illustrating the underlying morphological content contributing to the aggregated activations.

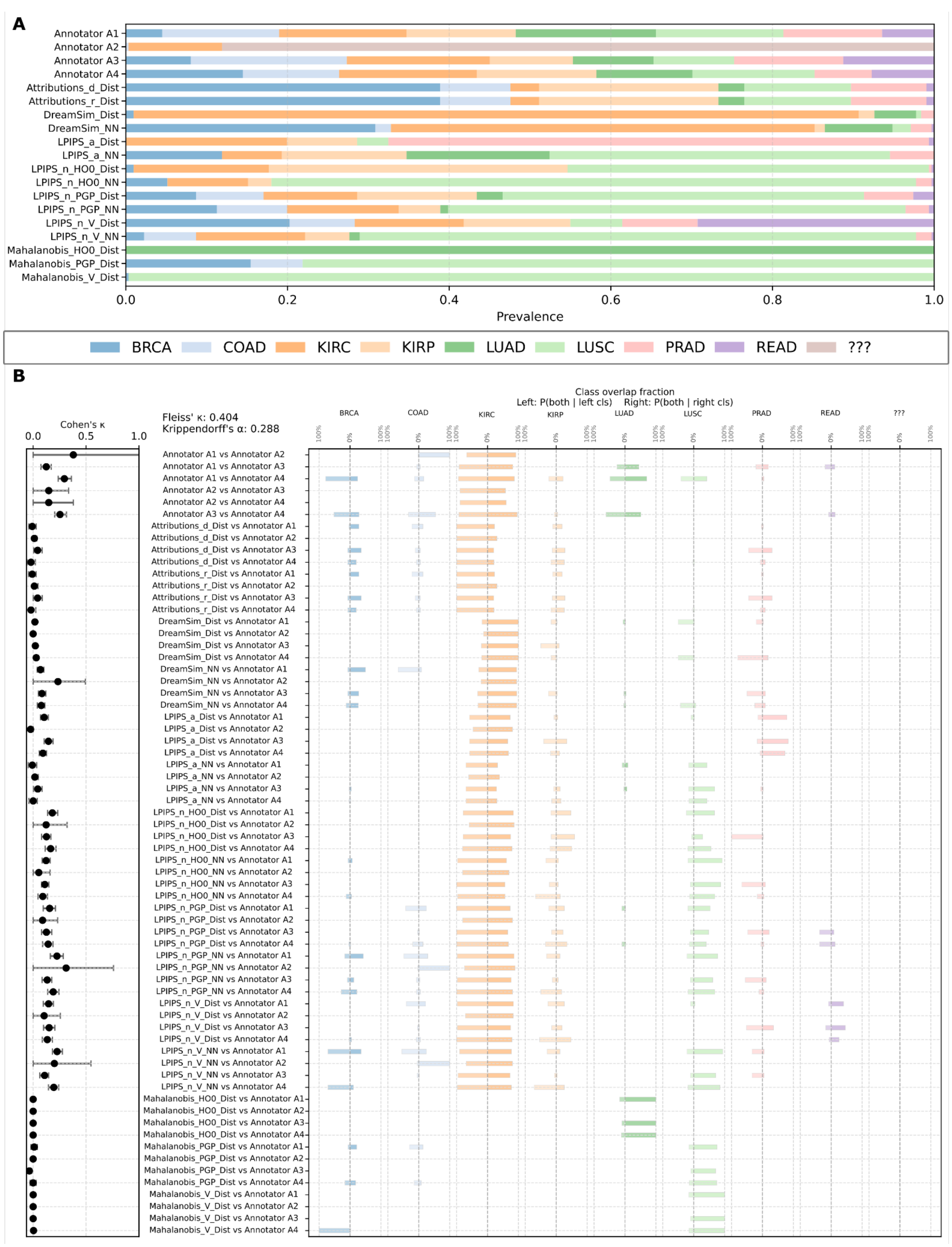

**Fig. S13 | Agreement and class coverage for activation atlas annotations in The Cancer Genome Atlas (TCGA-8 subclass) cancer task.**

Agreement analysis and class coverage for activation atlas cell annotations derived from the TCGA-8 subclass-level classification task.

**A**. Class coverage. Stacked bar plots showing the distribution of class assignments across activation atlas cells for each pathologist (Annotators A1–A4) and for each metric-based label assignment method. Bars indicate the proportion of atlas cells assigned to each TCGA-8 cancer subclass: breast invasive carcinoma (BRCA), colon adenocarcinoma (COAD), kidney renal clear cell carcinoma (KIRC), kidney renal papillary cell carcinoma (KIRP), lung adenocarcinoma (LUAD), lung squamous cell carcinoma (LUSC), prostate adenocarcinoma (PRAD), and rectum adenocarcinoma (READ). Where present, uncertain annotations ("???") are included as an explicit category to reflect overall class coverage.

**B**. Agreement metrics and class-wise overlap. Pairwise agreement between annotators and between annotators and metric-based assignments, computed excluding uncertain labels ("???") to ensure comparability of categorical agreement. Left: Cohen's κ values summarizing pairwise agreement, with points indicating mean κ across activation atlas cells and horizontal error bars denoting uncertainty estimated by bootstrap resampling. Overall inter-annotator agreement is summarized by Fleiss' κ and Krippendorff's α (top). Right: Class-wise overlap fractions indicating, for each cancer subclass, the fraction of activation atlas cells that received the same label from both compared sources. Bars on the left show, for each reference subclass, how often atlas cells with that reference label are assigned the same subclass by the comparator. Bars on the right show, for each assigned subclass, how often the assigned atlas cells originate from the corresponding reference subclass. Together, these measures capture asymmetric agreement and subclass-specific ambiguity across expert annotations and quantitative surrogate metrics.

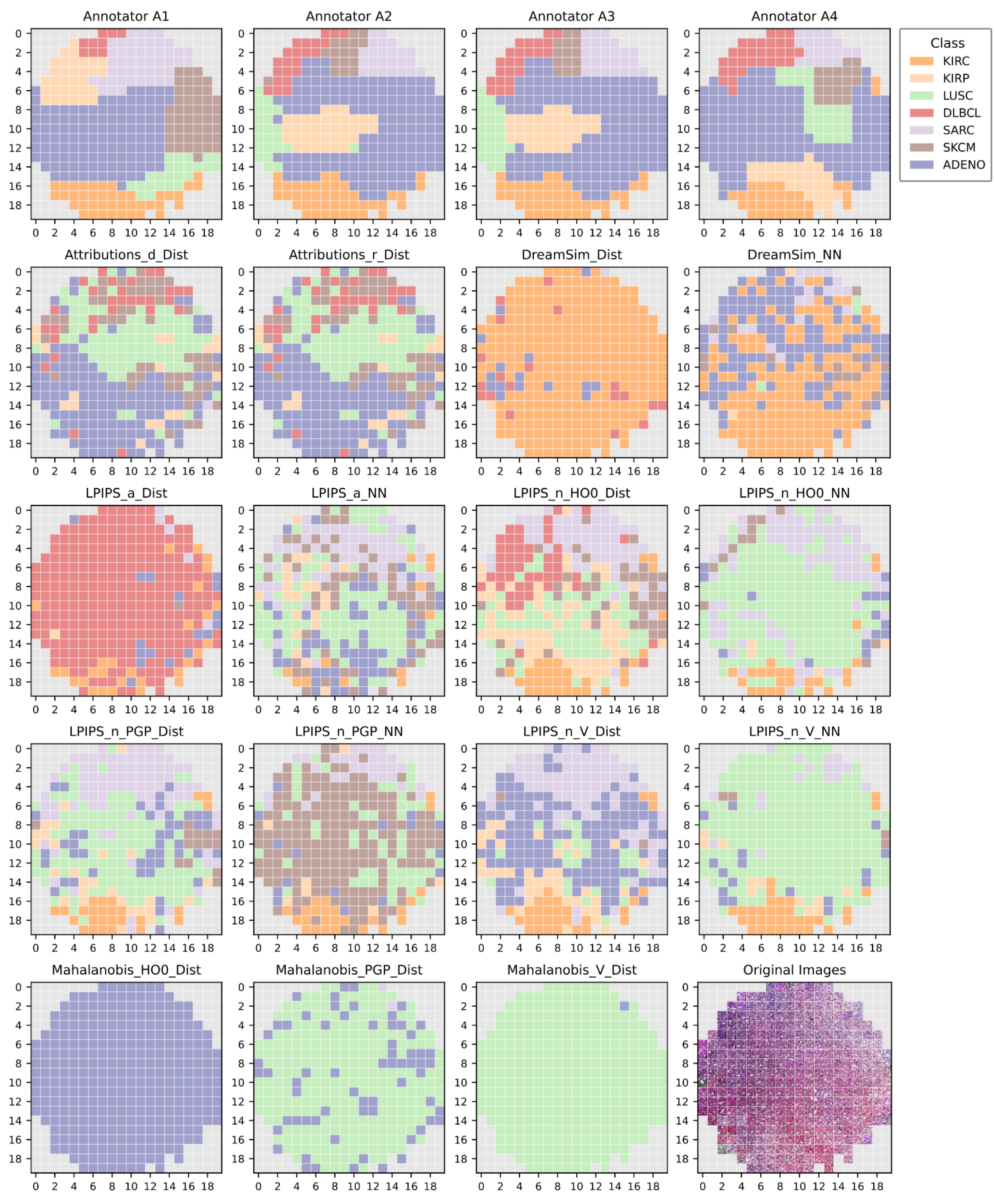

**Fig. S14 | Label maps for activation atlas cells in The Cancer Genome Atlas dataset (TCGA-11-adeno).**

Cell-wise label maps for an activation atlas derived from the TCGA-11 adenocarcinoma-level classification task. Each grid shows the same activation atlas layout, in which each cell corresponds to an aggregated region in the reduced activation embedding space. Cell colors indicate the class label assigned to each activation atlas cell.
Top row: Independent annotations by four pathologists (Annotators A1–A4), assigning one of the six TCGA cancer classes: adenocarcinoma (ADENO), kidney renal clear cell carcinoma (KIRC), kidney renal papillary cell carcinoma (KIRP), lung squamous cell carcinoma (LUSC), lymphoid neoplasm diffuse large B-cell lymphoma (DLBCL), sarcoma (SARC), and skin cutaneous melanoma (SKCM).
Middle rows: Corresponding atlas label maps obtained using quantitative surrogate assignment methods. Attribution-based methods assign labels based on class attributions aggregated over atlas cells, using either distance-based aggregation (_Dist; assignment based on minimal attribution distance) or nearest-neighbor-based aggregation (_NN; assignment based on the closest individual activations). Perceptual similarity-based methods assign each atlas cell to the closest class using either DreamSim or Learned Perceptual Image Patch Similarity (LPIPS), again using distance-based (_Dist) or nearest-neighbor (_NN) strategies. LPIPS variants differ by the feature extractor used, indicated by suffixes _a (AlexNet), _H00 (H-optimus-0), _PGP (Prov-GigaPath), and _V (vision transformer, UNI). Mahalanobis distance-based methods assign labels based on class-conditional feature distributions computed in the corresponding representation spaces.
Bottom right: Generated image patches associated with each activation atlas cell, illustrating the underlying morphological content contributing to the aggregated activations.

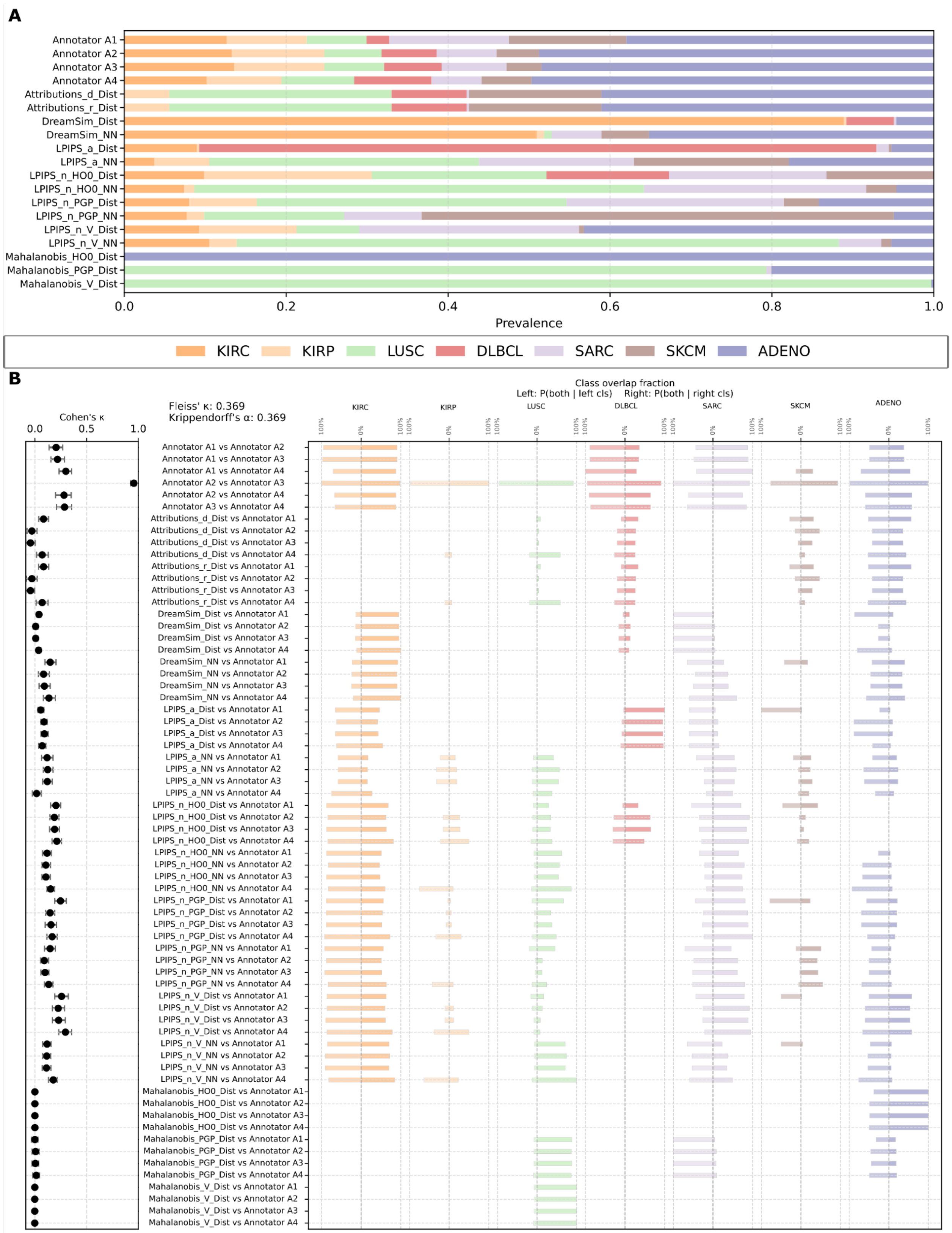

**Fig. S15 | Agreement and class coverage for activation atlas annotations in The Cancer Genome Atlas (TCGA-11 adeno) cancer task.**

Agreement analysis and class coverage for activation atlas cell annotations derived from the TCGA-11 adenocarcinoma-level classification task.

**A**. Class coverage. Stacked bar plots showing the distribution of class assignments across activation atlas cells for each pathologist (Annotators A1–A4) and for each metric-based label assignment method. Bars indicate the proportion of atlas cells assigned to each TCGA-11 cancer class: kidney renal clear cell carcinoma (KIRC), kidney renal papillary cell carcinoma (KIRP), lung squamous cell carcinoma (LUSC), lymphoid neoplasm diffuse large B-cell lymphoma (DLBCL), sarcoma (SARC), skin cutaneous melanoma (SKCM), and adenocarcinoma (ADENO).

**B**. Agreement metrics and class-wise overlap. Pairwise agreement between annotators and between annotators and metric-based assignments, computed excluding uncertain labels to ensure comparability of categorical agreement. Left: Cohen's κ values summarizing pairwise agreement, with points indicating mean κ across activation atlas cells and horizontal error bars denoting uncertainty estimated by bootstrap resampling. Overall inter-annotator agreement is summarized by Fleiss' κ and Krippendorff's α (top). Right: Class-wise overlap fractions indicating, for each cancer class, the fraction of activation atlas cells that received the same label from both compared sources. Bars on the left show, for each reference class, how often atlas cells with that reference label are assigned the same class by the comparator. Bars on the right show, for each assigned class, how often the assigned atlas cells originate from the corresponding reference class. Together, these measures capture asymmetric agreement and class-specific ambiguity across expert annotations and quantitative surrogate metrics.

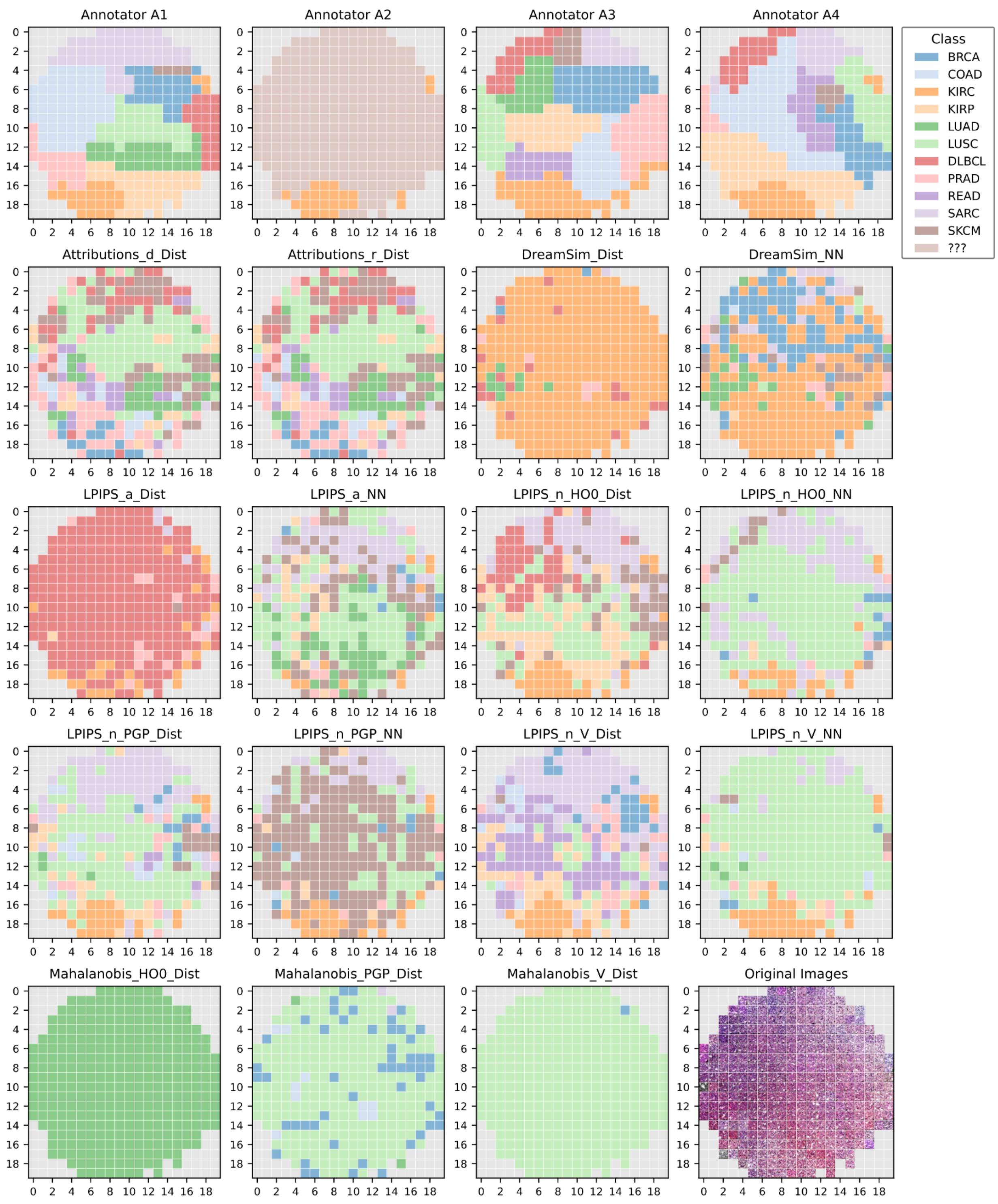

**Fig. S16 | Label maps for activation atlas cells in The Cancer Genome Atlas dataset (TCGA-11-subclass).**

Cell-wise label maps for an activation atlas derived from the TCGA-11 subclass-level classification task. Each grid shows the same activation atlas layout, in which each cell corresponds to an aggregated region in the reduced activation embedding space. Cell colors indicate the class label assigned to each activation atlas cell.
Top row: Independent annotations by four pathologists (Annotators A1–A4), assigning one of the eleven TCGA cancer subclasses: breast invasive carcinoma (BRCA), colon adenocarcinoma (COAD), kidney renal clear cell carcinoma (KIRC), kidney renal papillary cell carcinoma (KIRP), lung adenocarcinoma (LUAD), lung squamous cell carcinoma (LUSC), lymphoid neoplasm diffuse large B-cell lymphoma (DLBCL), prostate adenocarcinoma (PRAD), rectum adenocarcinoma (READ), sarcoma (SARC), and skin cutaneous melanoma (SKCM), or an explicit label for uncertainty (“???”).
Middle rows: Corresponding atlas label maps obtained using quantitative surrogate assignment methods. Attribution-based methods assign labels based on class attributions aggregated over atlas cells, using either distance-based aggregation (_Dist; assignment based on minimal attribution distance) or nearest-neighbor-based aggregation (_NN; assignment based on the closest individual activations). Perceptual similarity-based methods assign each atlas cell to the closest class using either DreamSim or Learned Perceptual Image Patch Similarity (LPIPS), again using distance-based (_Dist) or nearest-neighbor (_NN) strategies. LPIPS variants differ by the feature extractor used, indicated by suffixes _a (AlexNet), _H00 (H-optimus-0), _PGP (Prov-GigaPath), and _V (vision transformer, UNI). Mahalanobis distance-based methods assign labels based on class-conditional feature distributions computed in the corresponding representation spaces.
Bottom right: Generated image patches associated with each activation atlas cell, illustrating the underlying morphological content contributing to the aggregated activations.

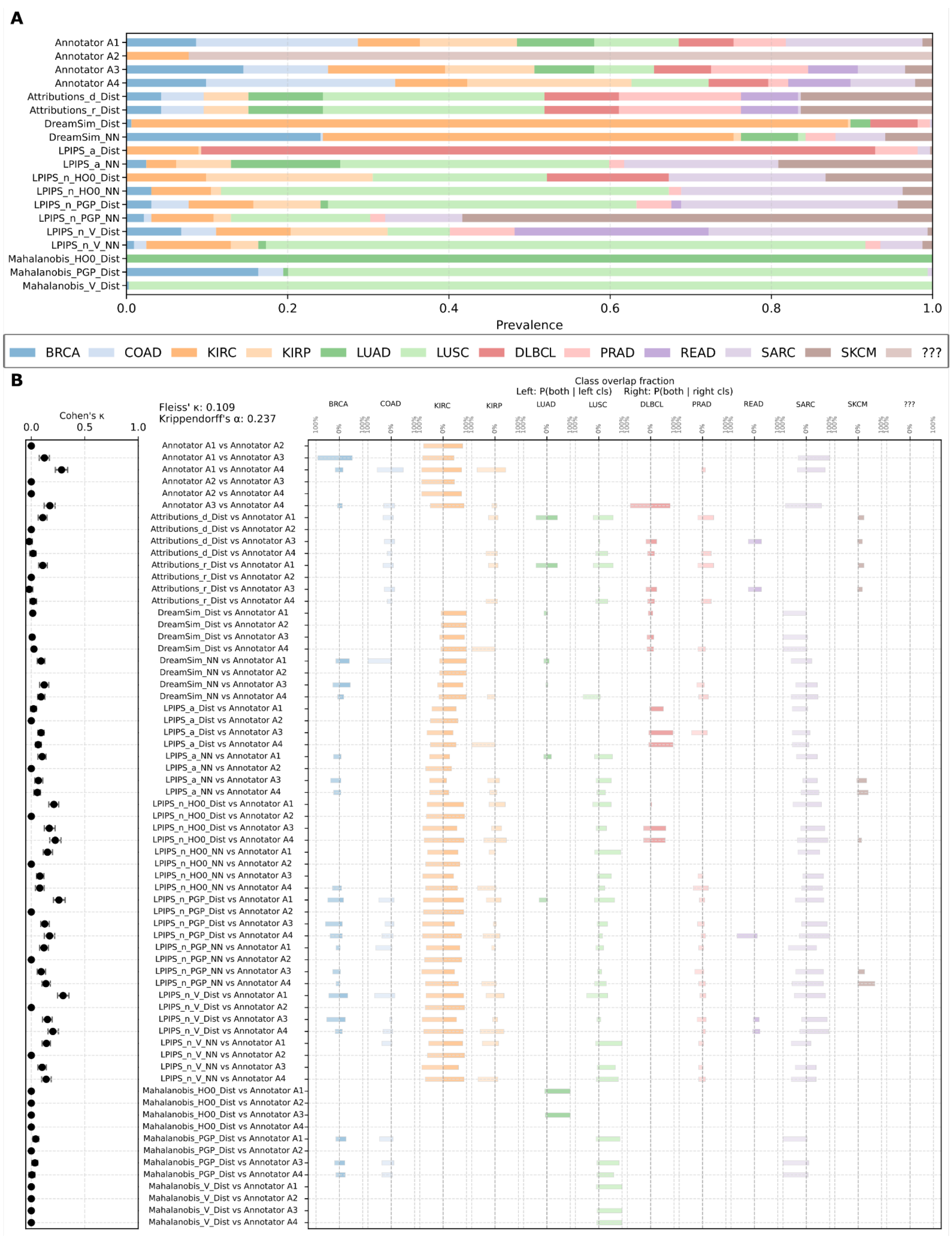

**Fig. S17 | Agreement and class coverage for activation atlas annotations in The Cancer Genome Atlas (TCGA-11 subclass) cancer task.**

Agreement analysis and class coverage for activation atlas cell annotations derived from the TCGA-11 subclass-level classification task.

**A**. Class coverage. Stacked bar plots showing the distribution of class assignments across activation atlas cells for each pathologist (Annotators A1–A4) and for each metric-based label assignment method. Bars indicate the proportion of atlas cells assigned to each TCGA-11 cancer subclass: breast invasive carcinoma (BRCA), colon adenocarcinoma (COAD), kidney renal clear cell carcinoma (KIRC), kidney renal papillary cell carcinoma (KIRP), lung adenocarcinoma (LUAD), lung squamous cell carcinoma (LUSC), lymphoid neoplasm diffuse large B-cell lymphoma (DLBCL), prostate adenocarcinoma (PRAD), rectum adenocarcinoma (READ), sarcoma (SARC), and skin cutaneous melanoma (SKCM). Where present, uncertain annotations ("???") are included as an explicit category to reflect overall class coverage.

**B**. Agreement metrics and class-wise overlap. Pairwise agreement between annotators and between annotators and metric-based assignments, computed excluding uncertain labels ("???") to ensure comparability of categorical agreement. Left: Cohen's κ values summarizing pairwise agreement, with points indicating mean κ across activation atlas cells and horizontal error bars denoting uncertainty estimated by bootstrap resampling. Overall inter-annotator agreement is summarized by Fleiss' κ and Krippendorff's α (top). Right: Class-wise overlap fractions indicating, for each cancer subclass, the fraction of activation atlas cells that received the same label from both compared sources. Bars on the left show, for each reference subclass, how often atlas cells with that reference label are assigned the same subclass by the comparator. Bars on the right show, for each assigned subclass, how often the assigned atlas cells originate from the corresponding reference subclass. Together, these measures capture asymmetric agreement and subclass-specific ambiguity across expert annotations and quantitative surrogate metrics.